\documentclass[10pt,twocolumn,letterpaper]{article}
\usepackage[pagenumbers]{cvpr} 

\usepackage[dvipsnames]{xcolor}

\usepackage{xcolor}
\definecolor{ered}{rgb}{0.72, 0.16, 0.2}

\newcommand{\q}[1]{``#1''} 

\newcommand{\std}[1]{\textbf{\scriptsize\textcolor{gray}{$\pm#1$}}} 

\RequirePackage{algorithm}
\RequirePackage{algorithmic}
\definecolor{cvprblue}{rgb}{0.21,0.49,0.74}
\usepackage[pagebackref,breaklinks,colorlinks,citecolor=cvprblue]{hyperref}
\usepackage{microtype}
\usepackage{graphicx}
\usepackage{booktabs} 
\usepackage{multirow}
\usepackage{bm}
\usepackage{enumitem}
\usepackage{hyperref}

\usepackage{listings}

\usepackage{amsmath}
\usepackage{amssymb}
\usepackage{mathtools}
\usepackage{amsthm}

\theoremstyle{plain}

\theoremstyle{definition}

\theoremstyle{remark}

\title{Recovering the Pre-Fine-Tuning Weights of Generative Models}

\author{Eliahu Horwitz \qquad Jonathan Kahana \qquad Yedid Hoshen\\
  School of Computer Science and Engineering\\
  The Hebrew University of Jerusalem, Israel\\
      \small\url{https://vision.huji.ac.il/spectral_detuning/}\\
  \small{\texttt{\{eliahu.horwitz, jonathan.kahana, yedid.hoshen\}@mail.huji.ac.il}}\\
}

\begin{document}
\maketitle
\begin{abstract}
    The dominant paradigm in generative modeling consists of two steps: i) pre-training on a large-scale but unsafe dataset, ii) aligning the pre-trained model with human values via fine-tuning. This practice is considered safe, as no current method can recover the unsafe, \textit{pre-fine-tuning} model weights. In this paper, we demonstrate that this assumption is often false. Concretely, we present \textit{Spectral DeTuning}, a method that can recover the weights of the pre-fine-tuning model using a few low-rank (LoRA) fine-tuned models. In contrast to previous attacks that attempt to recover pre-fine-tuning capabilities, our method aims to recover the exact pre-fine-tuning weights. Our approach exploits this new vulnerability against large-scale models such as a personalized Stable Diffusion and an aligned Mistral. 
\end{abstract}

\section{Introduction}
\label{sec:intro}
A key paradigm in deep learning is to first pre-train a foundation model \citep{llama2,code_llama} on a large, general-purpose dataset and then fine-tune the model for a specific task. Fine-tuning is used for critical applications including model safety \citep{red_teaming}, alignment to human preferences and values \citep{instructgpt,rlhf,dpo}, providing privacy guarantees \citep{lora_dp}, personalization \citep{dreambooth}, and more \citep{weak_to_strong,controlnet}. In this paper, we identify a vulnerability in fine-tuned models, wherein the pre-fine-tuning (Pre-FT) weights, i.e., the model weights before the fine-tuning stage, can be recovered using a small number of models fine-tuned via low-rank adaptation (LoRA) \cite{lora}.

To illustrate our setting, let us consider a Large Language Model (LLM). While the pre-trained version of the LLM exhibits advanced language understanding and generation capabilities, it is unaligned with human preference and is often deemed unsafe \citep{instructgpt,llama2}. These unsafe models can be used for example to get instructions for building a bomb or other malicious activities. To improve instruction following and enhance safety, model creators perform an alignment fine-tuning stage. Usually, only the aligned version of the LLM is published, and the recovery of the original Pre-FT unsafe weights, is implicitly assumed to be impossible. While for existing models the recovery of the Pre-FT weights poses a security and safety vulnerability; for future superhuman models, it may lead to catastrophic consequences.

\begin{figure*}[t]
\includegraphics[width=1.0\linewidth]{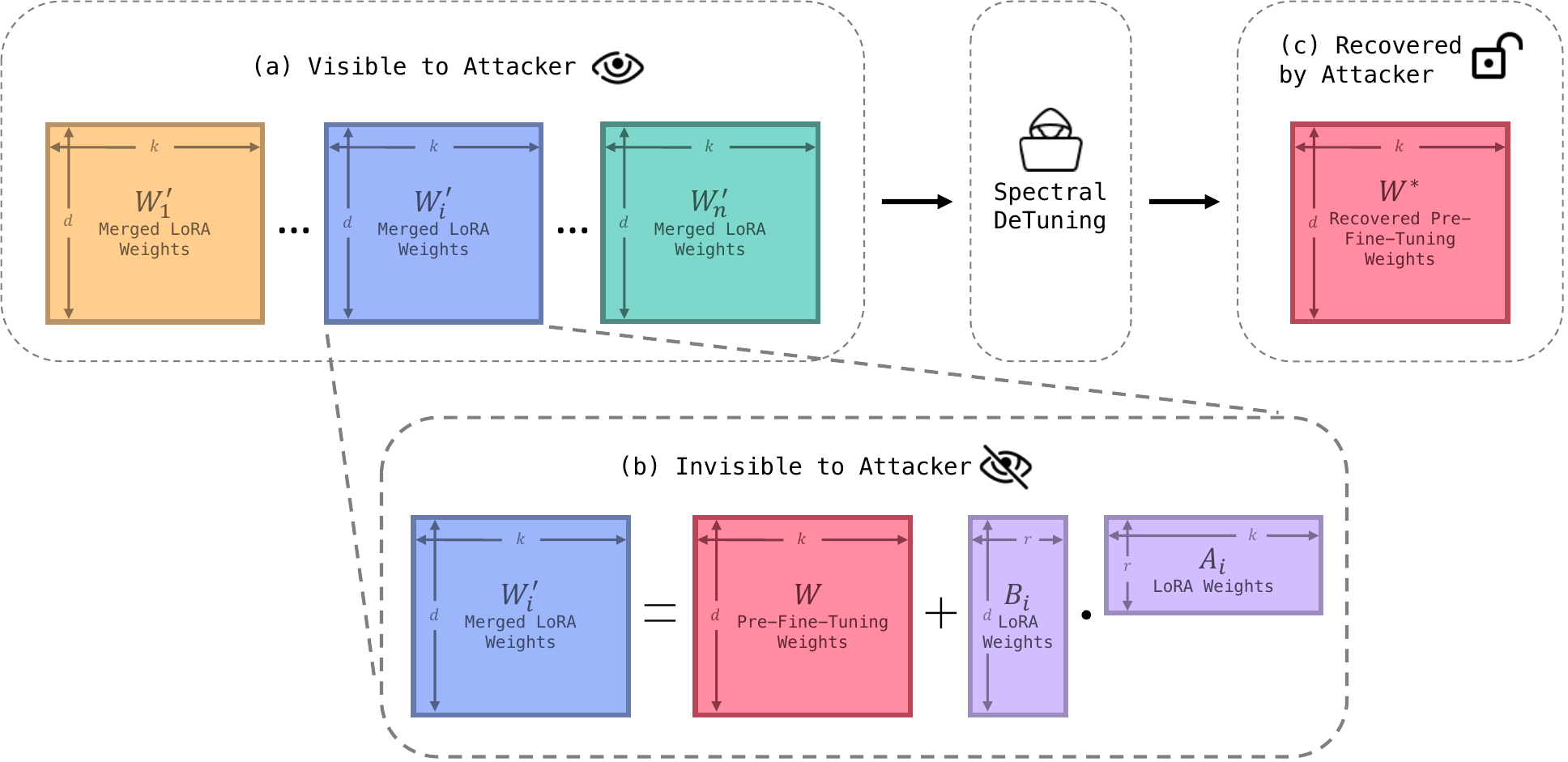}
 \caption{\textit{\textbf{Pre-Fine-Tuning Weight Recovery Attack Setting:}} We uncover a vulnerability in LoRA fine-tuned models wherein an attacker is able to undo the fine-tuning process and recover the weights of the original pre-trained model. The setting for the vulnerability is as follows: (a) The attacker only has access to $n$ different LoRA fine-tuned models. (b) The attacker assumes that all $n$ models originated from the same source model. \textbf{Note: The attacker has no access to the low-rank decomposition of the fine-tuned models.} (c) Using only the $n$ visible models, the attacker attempts to recover the original source model. Our method, \textit{Spectral DeTuning}, can perform the attack in an unsupervised and data-free manner on real models such as Stable Diffusion and Mistral. For simplicity, we illustrate the attack on a single layer, in reality, the attack is carried out independently on all the fine-tuned layers. Best viewed in color}
\label{fig:architectural_diagram}
\end{figure*}

Motivated by the above, we propose the task of \textit{Pre-Fine-Tuning Weight Recovery}. In this paper, we tackle this task in cases where multiple LoRA fine-tuned flavors of the same source model are available. We present an overview of our setting in \cref{fig:architectural_diagram}. This task is particularly timely due to two trends: i) Popular foundation models come in multiple flavors. E.g., LLaMA 2, Code LLaMA, Code LLaMA-Python, Code LLaMA-Instruct. ii) LoRA is becoming a key component for creating SoTA models \citep{mala500,perl}. These two trends have not yet merged, i.e, we are not aware of multi-flavored foundational models that use LoRA alignment fine-tuning. Here, we bring to the attention of the community the risks and perils involved in merging these trends. 
 
We present \textit{Spectral DeTuning}, a method that recovers the Pre-FT weights with remarkably high precision using iterative low-rank matrix factorization. To enhance optimization stability and accelerate convergence, we introduce a \textit{rank scheduler} that progressively increases the rank of the factorized matrices during optimization. A key distinction from prior attacks on model alignment \citep{adversarially_aligned,jailbroken,zou2023universal} is that Spectral DeTuning prioritizes restoring the exact Pre-FT \textit{weights} over Pre-FT \textit{functionalities}. It also does not require running inference through the model. This is advantageous as i) it does not require training data ii) it is highly parallelizable, e.g., on a cluster of desktop GPUs such as RTX2080 our method can recover the Pre-FT weights of a Mistral-7B model in under five minutes.

\begin{figure*}[t]
\includegraphics[width=1.0\linewidth]{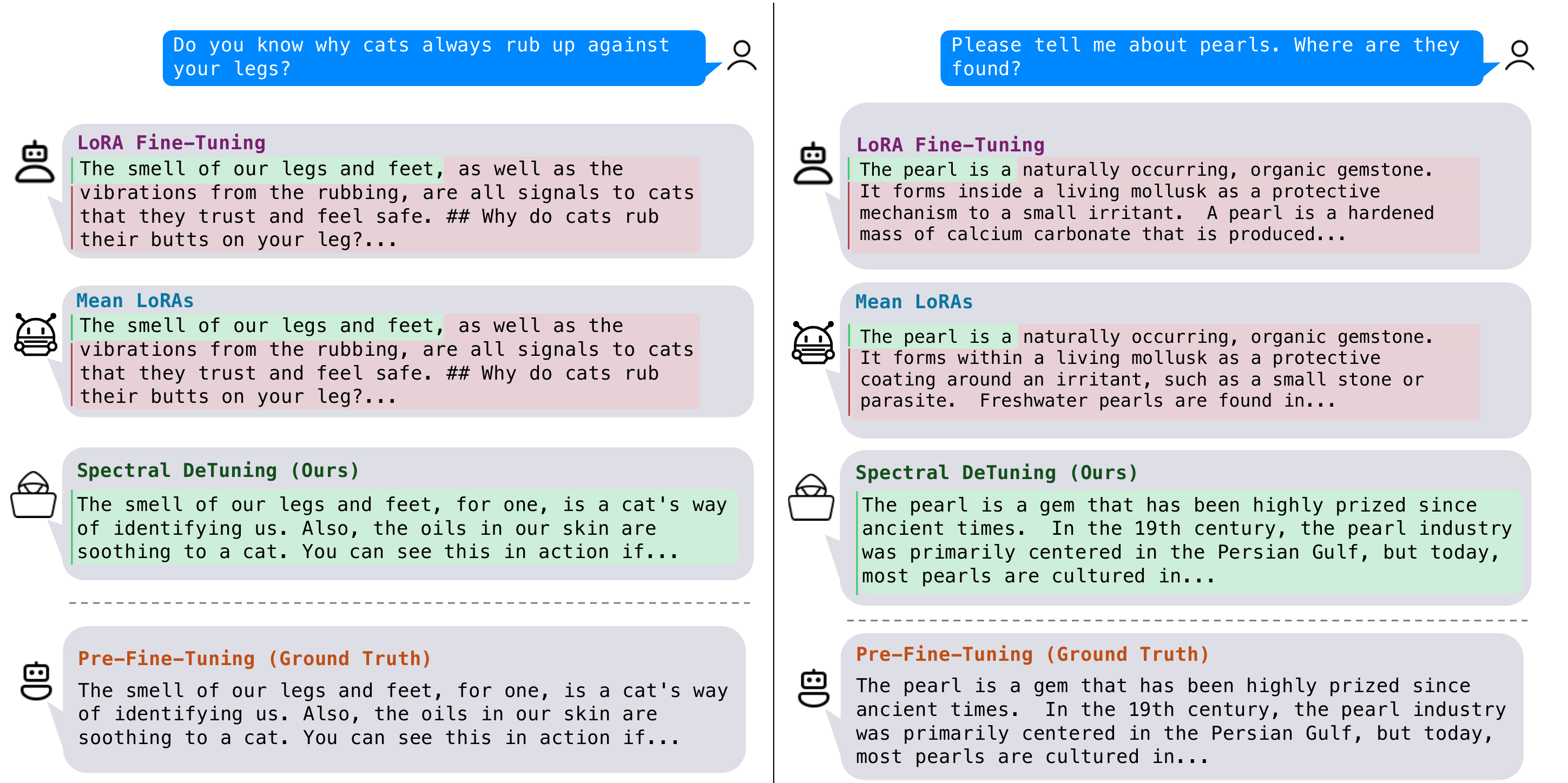}
\caption{\textit{\textbf{Mistral DPO Results:}} Our method, Spectral DeTuning, recovers the pre-fine-tuning generation capabilities with high precision, essentially undoing the DPO alignment LoRA fine-tuning. In green exact recovery, in red unrecovered words. Best viewed in color}
\label{fig:mistral_examples}
\end{figure*}

We demonstrate the effectiveness of our method by uncovering the vulnerability of real and widely used NLP and Vision models. Our approach achieves remarkable precision on an aligned Mistral model, effectively reversing the alignment training and restoring the original model (See \cref{fig:mistral_examples}). Similarly, on Stable-Diffusion, we recover the original model's weights with a vanishingly small error, showcasing almost perfect reconstruction of the original generation capabilities (See \cref{fig:sd_vis_rec}).

This work aims to stimulate research into preventing Pre-FT weight leakage and the associated risks in terms of model safety and alignment. To facilitate this research, we introduce \textit{LoWRA Bench}, a comprehensive benchmark comprising datasets and evaluation metrics, designed for assessing Pre-FT weight recovery methods.

To summarize, our main contributions are:
\begin{enumerate}
    \item Introducing the task of \textit{Pre-Fine-Tuning Weight Recovery}, a new attack vector against fine-tuned models. 
    \item Presenting \textit{Spectral DeTuning}, a highly effective method for pre-fine-tuning weight recovery attacks against state-of-the-art models.
    \item Providing \textit{LoWRA Bench}, a comprehensive suite of datasets and metrics designed for the evaluation of pre-fine-tuning weight recovery methods.  
\end{enumerate}

\section{Related Works}
\label{sec:related_works}
\subsection{Model Fine-tuning}
\label{sec:related_peft}
Model fine-tuning, crucial in deep learning research \citep{controlnet, lit, spa_text}, can be resource-intensive. Parameter-Efficient Fine-tuning (PEFT) methods \cite{lora,qlora,adapters,prefix_tuning,prompt_tuning,p_tuning,unified_peft,ia3,vpt,ada_lora,mpt,fedpara} aim to economize and broaden access to fine-tuning. These methods approximate full fine-tuning with fewer parameters. Some recent works combine multiple PEFT models \citep{ties_merging,mix_of_show,ZipLoRA,orthogonal_adaptation,lorahub}, hoping to leverage the strengths of individual models. LoRA \citep{lora} is perhaps the most popular PEFT method and is known for its effectiveness across various tasks and modalities \cite{prolific_dreamer,mplug,hyperdreambooth,break_a_scene}, sometimes even outperforming full fine-tuning. Given its popularity, in this paper, we focus on recovering Pre-FT weights of LoRA fine-tuned models.

\subsection{Model Safety and Security}
\label{sec:related_safety_privacy}
Deep learning models have various safety and security vulnerabilities. Membership inference attacks aim to detect if specific data samples were used in training \citep{mia,mia_a}. Model inversion attempts to generate the samples used during training \citep{model_inversion_images,model_inversion}. Machine unlearning protects against attacks by removing the effect of specific training samples without retraining the entire model \cite{machine_unlearning}. Model extraction, or model stealing, involves stealing a target model hidden behind an API by querying it multiple times \citep{model_stealing,beyond_labeling_oracles}. In contrast, Pre-FT weight recovery aims to recover the \textit{exact weights} of the pre-trained model, compromising the entire model rather than just a subset of capabilities. 
Additionally, our method, Spectral DeTuning, operates in an unsupervised and data-free manner.

\subsection{Model Red-Teaming and Adversarial Attacks}
\label{sec:related_red_teaming}
One of the primary methods for ensuring model safety involves incorporating human feedback through a reward model trained on annotator preferences, followed by reinforcement learning to fine-tune the model \citep{dpo, rlhf, red_teaming, red_teaming_hh, align_with_purpose, lora_rlhf}. However, \citet{alignment_limitations} argue that these alignment processes may leave undesired behavior partially intact and are thus vulnerable to adversarial prompting attacks. This has been demonstrated by red teaming \citep{red_teaming, red_teaming_hh} and adversarial attacks \citep{adversarially_aligned,jailbroken,zou2023universal} approaches. Unlike targeted attacks, Pre-FT weight recovery compromises the entire model by restoring the pre-trained weights. 
Moreover, our method, Spectral DeTuning, does not require running inference through the model.

\section{Preliminaries - LoRA}
\label{sec:preliminaries_lora}
Fine-tuning deep networks traditionally consisted of training all the network weights initialized by a pre-trained model. As this is costly for large-scale models, \citet{lora} recently introduced Low Rank Adaptation (LoRA). The authors postulate that the change in weights during fine-tuning often has a \q{low intrinsic rank}. They therefore introduced LoRA, which transforms each parameter matrix by the addition of a low-rank matrix. To create this low-rank matrix they multiply two full-rank matrices with suitable dimensions. This reparametrization drastically reduces the number of parameters being optimized. Specifically, for a pre-trained weight matrix $W_{\mathcal{P}}\in \mathbb{R}^{d\times k}$, the update $\Delta W$ can be decomposed into a rank $r$ decomposition $\Delta W = BA$ where $B \in \mathbb{R}^{d\times r}, A \in \mathbb{R}^{r\times k}$ and $r \ll min(d,k)$. During fine-tuning, $W_{\mathcal{P}}$ is frozen and only $A$ and $B$ are fine-tuned. This results in the following forward pass $W_{\mathcal{P}}x + \Delta W x = W_{\mathcal{P}}x + BAx $, where $x$ is the outcome of the previous layer. Since LoRA is linear by design, it is possible to merge the fine-tuned matrices back into the original matrix
\begin{equation} \label{eq:lora_weight_merge}
    W^{\prime}=W_{\mathcal{P}}+BA
\end{equation}
, thus introducing no additional parameters or inference latency to the original model. Originally, LoRA was applied to the query and value layers of attention blocks; however, it has been demonstrated that LoRA can be effectively extended to additional layers. Once merged, current models implicitly assume that recovering $W_{\mathcal{P}}$ and $BA$ from $W^{\prime}$ is impossible. \textbf{Throughout the paper, whenever we refer to the weights of a LoRA fine-tuned model, we assume the weights have been merged back as seen in \cref{eq:lora_weight_merge}.}

\section{Problem Definition}
\label{sec:problem_definition}
We introduce the task of \textit{Pre-Fine-Tuning Weight Recovery}. Its goal is to recover the Pre-FT weights of a given model, i.e., the weights of the original, pre-trained model. Specifically, in this work we assume that the fine-tuning was performed using LoRA.

\textit{Notation.} Formally, consider a model $\mathcal{F}_{\mathcal{P}}$ with $m$ fine-tuned layers that were fine-tuned via a rank $r$ LoRA and originated from the source model $\mathcal{P}$. We denote the weight matrices of  $\mathcal{F}_{\mathcal{P}}$ by $\{W^{\prime(j)}\}_{j=1}^{m}$ and those of $\mathcal{P}$ by $\{W^{(j)}_{\mathcal{P}}\}_{j=1}^{m}$ where both $W^{\prime(j)}$ and $W^{(j)}_{\mathcal{P}}$ are $\in \mathbb{R}^{d\times k}$. \textbf{Throughout the paper we assume the attacker does \textit{not} have access to $\mathcal{P}$ (nor to its weights $\{W^{(j)}_{\mathcal{P}}\}_{j=1}^{m}$).}

\textit{Attack setting.} The attacker has access to the weights of $n$ different $\mathcal{F}_{\mathcal{P}}$ models, all LoRA fine-tuned from the same pre-trained source model $\mathcal{P}$. The attack succeeds with precision $\epsilon$ if the attacker can accurately recover the weights of the pre-trained source model $\mathcal{P}$ up to an $\epsilon$ precision. Formally, given $\left\{\{W_{i}^{\prime(j)}\}_{j=1}^{m}\right\}_{i=1}^{n}$, the attacker needs find $\{W^{*(j)}\}_{j=1}^{m}$ such that

\begin{equation} \label{eq:pre_ft_objective}
\sum_{j=1}^{m} \left\| W^{(j)}_{\mathcal{P}} - W^{*(j)}\right \| < \epsilon
\end{equation}

We present an overview of this setting in \cref{fig:architectural_diagram}.

\textit{Success criteria.} We measure the success of the attack by the distance between the recovered weights and the original weights, in addition, in \cref{sec:benchmark} we discuss a number of ways to measure the success of the attack semantically.

\section{Spectral DeTuning}
\label{sec:method}
We now describe our method for carrying out a Pre-FT weight recovery attack. We start by introducing our optimization objective, followed by our optimization method and finally, a rank scheduler that stabilizes the optimization and results in better convergence. For simplicity, assume for now that all $n$ LoRA fine-tuned models used the same rank $r$, and that the value of $r$ is known to the attacker, in \cref{sec:rank_scheduler,sec:rank_estimation} we relax these assumptions. For brevity, we omit the layer index superscript $(j)$ and perform the same optimization across all layers independently.

\subsection{Optimization Objective}
\label{sec:optimization_objective}
To recover the Pre-FT weights, we need to predict $W_{\mathcal{P}}$ given $n$ fine-tuned weight matrices $\left\{W_{i}^{\prime}\right\}_{i=1}^{n}$. Leveraging their difference of up to $r$ principal components, we formulate the task as an optimization problem, where each LoRA provides additional constraints on $W_{\mathcal{P}}$. Specifically, recall that according to \cref{eq:lora_weight_merge}, $W_{i}^{\prime}$ can be decomposed into $W \in \mathbb{R}^{d\times k}$ and a rank $r$ matrix which we will denote by $M_{i} \in \mathbb{R}^{d\times k}$. Taking into account all $n$ different LoRA weights, we define the following objective 

\begin{equation} \label{eq:spectral_detuning_optimization}
    arg\min_{\substack{W,M_i \\ 1\leq i\leq n}} \sum_{i=1}^n \left\| W_{i}^{\prime} - (W + M_{i})\right \|^{2}_{2}~~~s.t.~~~rank~M_{i} \leq r
\end{equation}

Where $W \in \mathbb{R}^{d\times k}$ is the matrix we are optimizing to estimate $W_{\mathcal{P}}$. Intuitively, the objective optimizes the decomposition of each fine-tuned weight matrix into a \textit{shared} weight matrix which is the approximated Pre-FT matrix and an independent \textit{low rank} residual matrix. 

This objective exhibits desirable properties for an attacker. First, it is training-free, meaning, it requires no data, nor does it make any assumptions with regards to the data used to train the model. Moreover, the optimization is performed on a per-layer basis, enabling high parallelization of the attack. Finally, the objective is unsupervised, allowing an attacker to recover a model even when they have no prior knowledge regarding the source model.

 \begin{figure}[t]
 \includegraphics[width=1.0\linewidth]{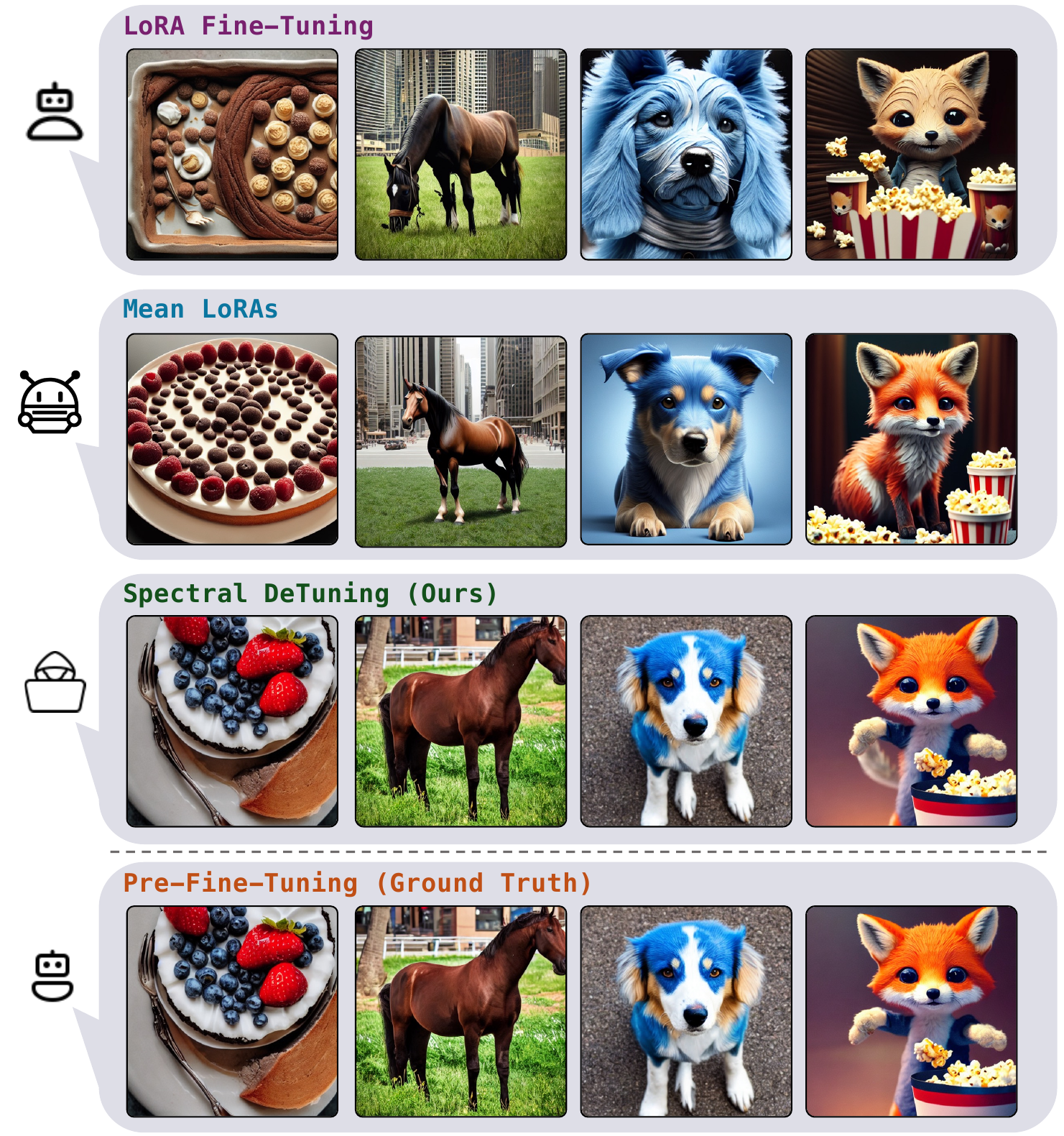}
 \caption{\textit{\textbf{Stable Diffusion Results:}} Spectral DeTuning recovers the Pre-Fine-Tuning images with high precision, even when using \q{in the wild} LoRAs, essentially reversing the personalization fine-tuning of the LoRA model}
\label{fig:sd_vis_rec}
\end{figure}

\subsection{Pre-FT Weight Recovery Algorithm}
\label{sec:pre_ft_methods}
We propose Spectral DeTuning, an iterative, gradient-free algorithm for Pre-FT weight recovery. The method is fast (even on CPU) and is easily parallelizable. The core idea is that while the optimization problem in \cref{eq:spectral_detuning_optimization} is non-convex, it can be iteratively broken down into a set of simple sub-problems which have closed-form solutions. Our procedure has three major components: initialization, M-step and W-step. Note, solving \cref{eq:spectral_detuning_optimization} requires optimizing $n+1$ matrices, i.e., $W$ and $M_1,M_2,...,M_n$.

\textit{Initialization.} At iteration $0$, we set $W^*$ as the average of all the fine-tuned matrices, i.e., $W^{*}=\frac{1}{n} \sum_{i=1}^n W_{i}^{\prime}$.

\textit{M-step.} We solve the optimization problem by coordinate descent \cite{coordinate_decent}. We first fix $W^{*}$ and solve for $\{M_{i}\}_{i=1}^{n}$. Note that when $W^*$ is given, the optimization problems for each $M_1,..,M_n$ are decoupled.  Specifically, at each iteration, the optimization problem for $M_{i}$ is: 

\begin{equation} \label{eq:m_update}
M_i^{*} = arg\min_{M_i} \|(W_i^{\prime} - W^*) - M_i\|_2^2~~~s.t.~~~rank~M_{i} \leq r
\end{equation}

Luckily, the solution to this optimization problem is available in closed-form and is given by the \q{Singular Value Decomposition} (SVD) of $W_i^{\prime} - W^*$.
The optimal value of $M_i$ is:

\begin{gather} \label{eq:m_trunc}
U_{i}, \Sigma_{i}, V^{T}_{i} = \text{SVD}(W_i^{\prime} - W^*) \\
M^*_i=U_{i} \Sigma_{i|r} V^{T}_{i} \nonumber
\end{gather}

Where $\Sigma_{i|r}$ represents the top $r$ singular values of $\Sigma_{i}$.

\textit{W-step.} By fixing the values of $M^*_1,..,M^*_n$, we can easily compute the optimal value of $W$. The optimization problem is given by:

\begin{equation} \label{eq:w_update_opt}
W^* = arg\min_{W} \sum_{i=1}^n \|(W_i^{\prime} - M^*_i) - W\|_2^2
\end{equation}

By simple calculus, the closed-form solution is:

\begin{equation} \label{eq:w_update}
W^* = \frac{1}{n} \sum_{i=1}^n \left(W_i^{\prime} - M^*_i\right)
\end{equation}

We iterate between the M-step and W-step until convergence. As shown in \cref{alg:spectral_detuning}, the algorithm can be easily implemented in as little as $8$ lines of python.

\subsection{Rank Scheduler}
\label{sec:rank_scheduler}
The algorithm proposed in \cref{sec:pre_ft_methods} tends to perform well in general.
However, we empirically found that solving the optimization problem with high ranks can result in slow and inaccurate convergence. We therefore introduce a rank scheduler. The idea of the rank scheduler is to start by forcing $M_{i}$ to be of rank $r^{*}<r$, allowing Spectral DeTuning to focus on the most significant principal components first. $r^{*}$ is increased according to a schedule until finally $r^{*}=r$. Specifically, we use an \q{Increase on Plateau} type of scheduler where the rank is increased whenever the loss term from  \cref{eq:spectral_detuning_optimization} plateaus.
When not all LoRAs have the same rank, we assign a distinct rank scheduler to each LoRA. The rank scheduler requires knowing the LoRA rank; we show how to estimate it in \cref{sec:rank_estimation}. For more details see \cref{app:impl_details}. We show empirically in \cref{sec:rank_ablation} that there are cases where the rank scheduler improves the rate and quality of convergence significantly.

\begin{algorithm}[t]
   \caption{PyTorch Pseudocode for Spectral DeTuning}
   
   \label{alg:spectral_detuning}
   
    \definecolor{codeblue}{rgb}{0.25,0.5,0.5}
    \definecolor{codekw}{rgb}{0.85, 0.18, 0.50}
    \lstset{
  backgroundcolor=\color{white},
  basicstyle=\fontsize{7.1pt}{7.1pt}\ttfamily\selectfont,
  columns=fullflexible,
  breaklines=true,
  captionpos=b,
  commentstyle=\fontsize{7.1pt}{7.1pt}\color{codeblue},
  keywordstyle=\fontsize{7.1pt}{7.1pt}\color{codekw},
}
\begin{lstlisting}[language=python]
# W_ps: List of n fine-tuned weight matrices
# steps: Number of optimization steps
# r: LoRA rank

# Initialize W_star
W_s = torch.mean(torch.stack(W_ps), axis=0)

# Perform optimization
for step in range(steps):
    # M-step
    # Approximate each M^*_i (Eq. 5)
    M_s = [W_p - W_s for W_p in W_ps]

    # Truncate each M^*_i to rank <= r (Eq. 5)
    for i in range(len(M_s)):
        (U, S, V) = torch.svd_lowrank(M_s[i], q=r)
        M_s[i] = (U @ torch.diag_embed(S)) @  V.T


    # W-step
    # Approximate W_star (Eq. 7)
    W_s = [W_p - M_si for (W_p, M_si) in zip(W_ps, M_s)]
    W_s = torch.mean(torch.stack(W_s), axis=0)
\end{lstlisting}
\end{algorithm}

\subsection{LoRA Rank Estimation}
\label{sec:rank_estimation}
We propose an effective heuristic for estimating LoRA rank.
Assume we have two LoRA fine-tuned models $W^{\prime}_i = W + M_i$ and $W^{\prime}_j = W + M_j$, where the ranks of $M_i,M_j$ are $r_i,r_j$ respectively. While it is not trivial to recover the rank of $M_i$ solely by observing $W^{\prime}_i$, there is a trick. Subtracting the two fine-tuned models obtains $W^{\prime}_i - W^{\prime}_j = M_i - M_j$. Importantly, the rank $W^{\prime}_i - W^{\prime}_j$ is upper bounded by $r_i + r_j$, i.e., $rank(W_i'-W_j') \leq r_i+r_j$. Given $n$ LoRAs, there are $\frac{n(n-1)}{2}$ distinct inequalities for the $n$ unknown ranks $r_1, r_2, .., r_n$.

We can formulate this as a linear programming problem as follows:
\[
\begin{array}{ll}
\underset{\mathbf{r}}{\text{minimize}} & \mathbf{1}^T \mathbf{r} \\
\text{subject to} & \mathbf{A} \mathbf{r} \geq \mathbf{b} \\
& r_i \geq 1, \quad \forall i
\end{array}
\]

\noindent where:
\begin{itemize}
    \item $\mathbf{A} \in \{0,1\}^{n^2,n}$ so that $A_{ni+j, i} = 1$ and $A_{ni+j, j} = 1$ and $0$ elsewhere.    
    \item $ \mathbf{b} \in R^{n^2} $ so that $ b_{ni+j} = rank(W^{\prime}_i - W^{\prime}_j)$.

\end{itemize}

\begin{figure*}[t!]
    \begin{center}
    \includegraphics[width=1.0\linewidth]{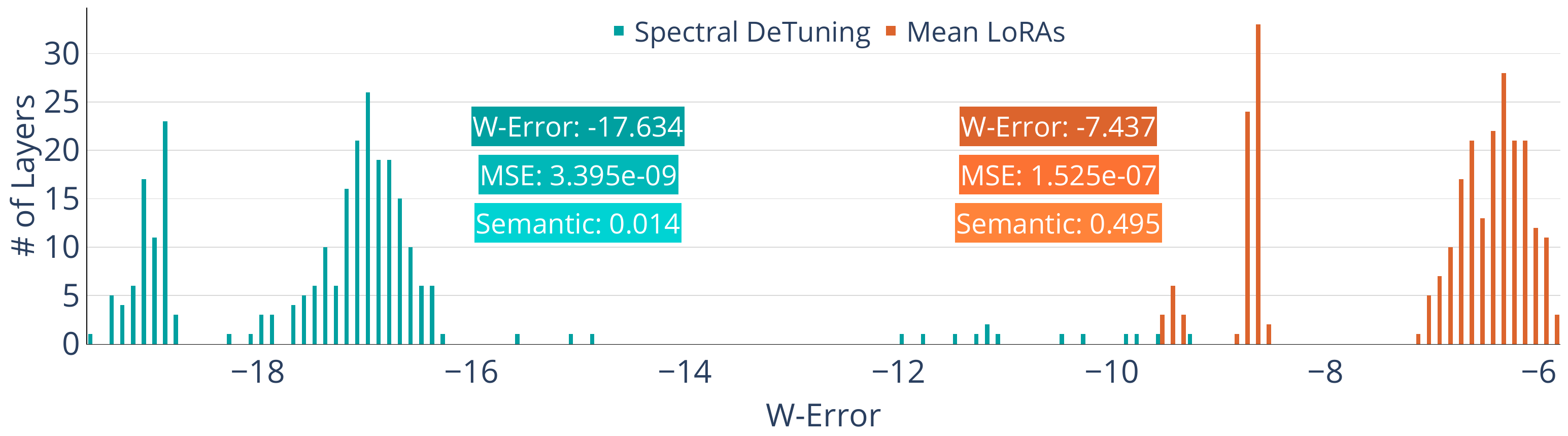}
    \end{center}
    \caption{\textit{\textbf{Motivation for the Log in W-Error:}} We visualize the convergence of all layers using Spectral DeTuning and the Mean LoRAs baselines. Spectral DeTuning clearly converges to a much better solution for almost all layers. Note that MSE does not summarize the convergence well as it yields the value of the poorly converging outlier layers. The W-Error better conveys the actual convergence by working in log-space. Results for a random subset of $5$ Stable Diffusion LoRAs}
    \label{fig:w_error_log_motivation}
\end{figure*}

In practice, we populate $b$ using a \textit{numerical rank} computed via the multiplicative gap following a similar protocol to \citep{carlini_stealing}. Using an off-the-shelf linear programming solver accurately retrieves the ranks. We demonstrate the accuracy of this method in  \cref{sec:rank_ablation}, the unknown ranks were recovered perfectly in all cases.

\section{LoWRA Bench}
\label{sec:benchmark}
We present \textit{\textbf{Lo}RA \textbf{W}eight \textbf{R}ecovery \textbf{A}ttack} (LoWRA) Bench, a comprehensive benchmark designed to evaluate Pre-FT weight recovery methods.

\subsection{Dataset}
\label{sec:dataset}
Our dataset encompasses three pre-trained representative source models: a Vision Transformer (ViT) \citep{vit} trained on ImageNet-1K \citep{imagenet}, Mistral-7B-v0.1 \cite{mistral}, and Stable Diffusion 1.5 \cite{latent_diffusion}. These models collectively cover supervised and self-supervised objectives, spanning both vision and natural language processing (NLP) domains, as well as generative and discriminative tasks. Notably, these models are widely used and deployed in numerous production systems. See \cref{table:pre_ft_bench} for an overview of the dataset.

For each source model, we curate $15$ LoRA models fine-tuned on diverse datasets, tasks, and objectives. The dataset comprises a diverse array of layer types, including self-attention, cross-attention, and MLPs. This diversity enables us to assess the generalization capabilities of Pre-FT methods. The evaluation can be conducted on a per-model basis, per layer type, or per layer depth, allowing for a comprehensive analysis of Pre-FT methods. Overall, our dataset includes $544$ source model layers. When taking into account the fine-tuned LoRA layers, the dataset includes over $8,000$ layers. For further details see \cref{app:lowra_bench}.

\subsection{Numeric Evaluation Metrics}
\label{sec:evaluation_metrics}
\noindent \textit{Weight Error (W-Error).} 
We measure numeric convergence by the mean squared weight error (as defined in \cref{eq:pre_ft_objective}) and average across all layers in log space:
\begin{equation} \label{eq:w_error}
    \frac{1}{m}\sum_{j=1}^m \left(\log_{10}\left(MSE(W^{(j)}_{\mathcal{P}} - W^{*(j)}\right)\right)
\end{equation}
We use log-space as when errors are very small, the average mean squared weight error is determined by outliers, e.g., a single non-converging layer when all other layers converge. Log transforming the mean squared error is robust to such outliers. We visualize this in \cref{fig:w_error_log_motivation} where Spectral DeTuning clearly converges to a much better solution. Despite the outstanding convergence, the small number of outliers create a false impression where the MSE shows a significantly higher error. In \cref{app:w_error_vs_lpips} we show that the W-Error is strongly correlated with the recovery of the Pre-FT semantic capabilities ($\rho=0.880$ for W-Error vs. LPIPS).

\begin{table}[t]
\caption{\textit{\textbf{LoWRA Bench Dataset Summary:}} The dataset covers widely used models spanning vision and language modalities. It includes over $540$ Pre-FT layers and over $8,000$ fine-tuned layers}
\begin{center}
    \begin{tabular}{l@{\hskip5pt}l@{\hskip5pt}l@{\hskip5pt}c@{\hskip5pt}c@{\hskip5pt}}    
        \begin{tabular}[l]{@{}l@{}}Pre-FT\\Model\end{tabular} & Task & Fine-tuning Task & \begin{tabular}[l]{@{}l@{}}\# Pre-FT\\Layers\end{tabular}  & \begin{tabular}[l]{@{}l@{}}\# FT \\ Layers\end{tabular} \\
        \midrule
       ViT          & Classific.       & VTAB-1K            & $24$ & $360$     \\
        SD1.5        & T2I Gen. & Personalization    & $264$ & $3960$   \\
       Mistral    & Text Gen.            & UltraChat SFT      & $128$ & $1920$   \\
       Mistral    & Text Gen. & UltraFeedback DPO  & $128$ & $1920$   \\
    \end{tabular}
\end{center}
\label{table:pre_ft_bench}
\end{table}

\subsection{Semantic Evaluation Metrics}
\label{sec:semantic_evaluation_metrics}
We design model specific metrics focusing on the Pre-FT task from a semantic perspective.

\noindent\textit{ViT Activation Distance (Act.-Dist.).} We take the cosine distance between the activations of the Pre-FT model and those of the recovered one. Specifically, we take the mean of all transformer tokens at the end of the last transformer block. We use a subset of $5000$ images from the ImageNet validation set.

\noindent \textit{Stable Diffusion LPIPS (LPIPS).} The LPIPS \citep{lpips} distance between images generated by the Pre-FT model and by the recovered model. We report the mean LPIPS for the first $100$ prompts of the COCO Captions validation dataset \cite{coco_captions}.

\noindent \textit{Mistral SBERT (SBERT).} The log cosine distance between the Sentence-BERT \citep{sbert} (SBERT) textual embeddings of text generated by the Pre-FT model and by the recovered model. We report the mean log cosine for the first $100$ prompts of the Alpaca Farm evaluation benchmark \cite{alpaca_farm}. 

\subsection{Experimental Setup}
\label{sec:experimental_setup}
\textit{Subsets.} In each experiment, we specify a number of LoRA fine-tuned models $L$, which is often lower than the total number of LoRAs available in the datasets. We do this by randomly sampling a set of $L$ models from the datasets. We then perform the Pre-FT weight recovery method on this subset. We repeat this experiment (including subset sampling) $10$ times. The reported performance metrics are the average and standard deviation over the experiments.

\textit{Baselines.} The two baseline methods are i) using one of the fine-tuned LoRA models; we average the results over all models in the sampled subset. ii) averaging the weights across all LoRA fine-tuned models in the sampled subset and reporting the results of the weight averaged model. 
The motivation behind the mean LoRA baseline, is the assumption that the mean of the residuals is the zero matrix, i.e., $\frac{1}{n}\sum_{i=1}^{n}M_{i}=0$. In this case the optimum of \cref{eq:spectral_detuning_optimization} becomes the average of all the weights.

\section{Experiments}
\label{sec:experiments}

\begin{table}[t]
    \caption{\textit{\textbf{ViT Results:}} As expected, the LoRA fine-tuned models have drifted away from the initial weights and activations. The mean of the LoRAs is slightly better, but is still far from the Pre-FT model. In contrast, Spectral DeTuning achieves an almost perfect semantic convergence. Reported results use $n=5$ fine-tuned LoRAs}
    \begin{center}
    \begin{tabular}{l@{\hskip5pt}c@{\hskip5pt}c@{\hskip5pt}}    
            Method  & W-Error  $\downarrow$ & Act.-Dist. $\downarrow$ \\
            \midrule
            LoRA FT         & -4.602 \std{0.110}  & 1e{-}1 \std{9e{-}2}\\
            Mean LoRA       & -5.214 \std{0.114}  & 5e{-}2 \std{1e{-}2}\\
            Spectral DeTuning    & \textbf{-15.942} \std{\textbf{1.889}} & \textbf{1e{-}6} \std{\textbf{3e{-}6}}\\
        \end{tabular}
        \end{center}
        \label{table:vit_results}
\end{table}

\subsection{Preliminary Investigation on ViT}
\label{sec:vit}
We begin our exploration of Pre-FT weight recovery using ViT, due to its simple architecture with consistent weight dimensions and relatively small model size. While this is our simplest task, it is not a \q{toy example} but a real model that is widely used and deployed in countless production settings. 
In \cref{table:vit_results} we show the results for $n=5$ fine-tuned LoRAs. As expected, the LoRA fine-tuned models are indeed different from the Pre-FT model. Averaging over several LoRA models slightly improves the results, but is still far from recovering the Pre-FT activations. Our method, Spectral DeTuning, performs much better and  attains an almost perfect semantic convergence, outperforming the baselines by a wide margin.

\subsection{In the Wild Weight Recovery of Stable Diffusion}
\label{sec:stable_diffusion}
Having shown the vulnerability of an image classification model, we now test the vulnerability of Stable Diffusion, a multi-modal text-to-image model. To this end, we used publicly fine-tuned LoRAs found on \textit{civitai}, allowing us to validate our method \q{in the wild}. As in the case of ViT, the baselines perform poorly on all metrics. In contrast, Spectral DeTuning recovers the Pre-FT weights with high precision. This results in a significant improvement of the recovered semantic capabilities of the Pre-FT model while using as little as $n=5$ fine-tuned LoRAs (See \cref{table:stable_diffusion_results} and \cref{fig:sd_vis_rec}). 

\textcolor{ered}{Implication: SoTA personalization methods using LoRA are vulnerable to Pre-FT weight recovery attacks.}

\subsection{Pre-FT Weight Recovery of an Aligned LLM}
\label{sec:mistral}
Having achieved success with mid-sized image models, we now investigate the ability of our method to scale up to a large-scale aligned LLM. Specifically, we use Mistral-7B, a top performing open-source LLM. Following common practice, we fine-tune the model in two stages, first performing supervised fine-tuning (SFT) followed by a direct preference optimization (DPO) alignment fine-tuning stage \citep{dpo}. We report the results of both stages in \cref{table:mistral_results}, as we can see, Spectral DeTuning successfully recovers the weights with high precision. This high quality recovery is also expressed in recovering the semantic capabilities of the Pre-FT model. I.e., the estimated weights yield a model which provides responses that are very similar to the Pre-FT model and much more so than the LoRA fine-tuned model (See \cref{fig:mistral_examples}).

\textcolor{ered}{Implication: SoTA LLMs that use LoRA for alignment fine-tuning are vulnerable to Pre-FT weight recovery attacks.}

\begin{table}[t]
    \caption{\textit{\textbf{Stable Diffusion Results:}} Spectral DeTuning is almost three times better than the baselines, recovering a large portion of the semantic capabilities of the pre-fine-tuning Stable Diffusion. Reported results use $n=5$ fine-tuned LoRAs taken from an online LoRA marketplace}
    \begin{center}
    \begin{tabular}{l@{\hskip5pt}c@{\hskip5pt}c@{\hskip5pt}}    
            Method  & W-Error $\downarrow$   & LPIPS $\downarrow$\\
            \midrule
            LoRA FT  & -6.921 \std{1.080}  & 0.514 \std{0.047} \\
            Mean LoRA  & -7.540 \std{1.099}  & 0.482 \std{0.012} \\
            Spectral DeTuning  & \textbf{-17.816} \std{\textbf{2.126}}  & \textbf{0.009} \std{\textbf{0.006}} \\
        \end{tabular}
        \end{center}
        \label{table:stable_diffusion_results}
\end{table}

\begin{table}[t]
\caption{\textit{\textbf{Mistral Results:}} Spectral DeTuning recovers the Pre-FT weights and semantic capabilities with high precision, both in the supervised fine-tuning (SFT) stage and the alignment fine-tuning stage (DPO). Reported results use $n=12$ fine-tuned LoRAs for SFT and $n=8$ fine-tuned LoRAs for DPO}
\begin{center}
    \begin{tabular}{c@{\hskip5pt}l@{\hskip5pt}c@{\hskip5pt}c@{\hskip5pt}}    
        & Method  & W-Error $\downarrow$ & SBERT $\downarrow$ \\
        \midrule
       \multirow{3}{*}{\rotatebox[origin=c]{90}{SFT}}  & LoRA FT  & -8.677 \std{0.153} & -0.994 \std{0.731}\\
        & Mean LoRA  & -9.299 \std{0.222} & -1.007 \std{0.726}\\
        & Spectral DeTuning  & \textbf{-16.502} \std{\textbf{1.855}} & \textbf{-9.324} \std{\textbf{6.942}} \\
        \cmidrule{2-4}
        
        \multirow{3}{*}{\rotatebox[origin=c]{90}{DPO}} & LoRA FT  & -9.903 \std{0.166} & -3.058 \std{4.763}\\
        & Mean LoRA  & -10.757 \std{0.178} & -3.455 \std{5.171}\\
        & Spectral DeTuning  & \textbf{-22.062} \std{\textbf{1.180}} & \textbf{-14.708} \std{\textbf{3.123}} \\
    \end{tabular}
\end{center}
\label{table:mistral_results}
\end{table}

\section{Ablations}
\label{sec:rank_ablation}
\begin{figure}[t]
 \includegraphics[width=1.0\linewidth]{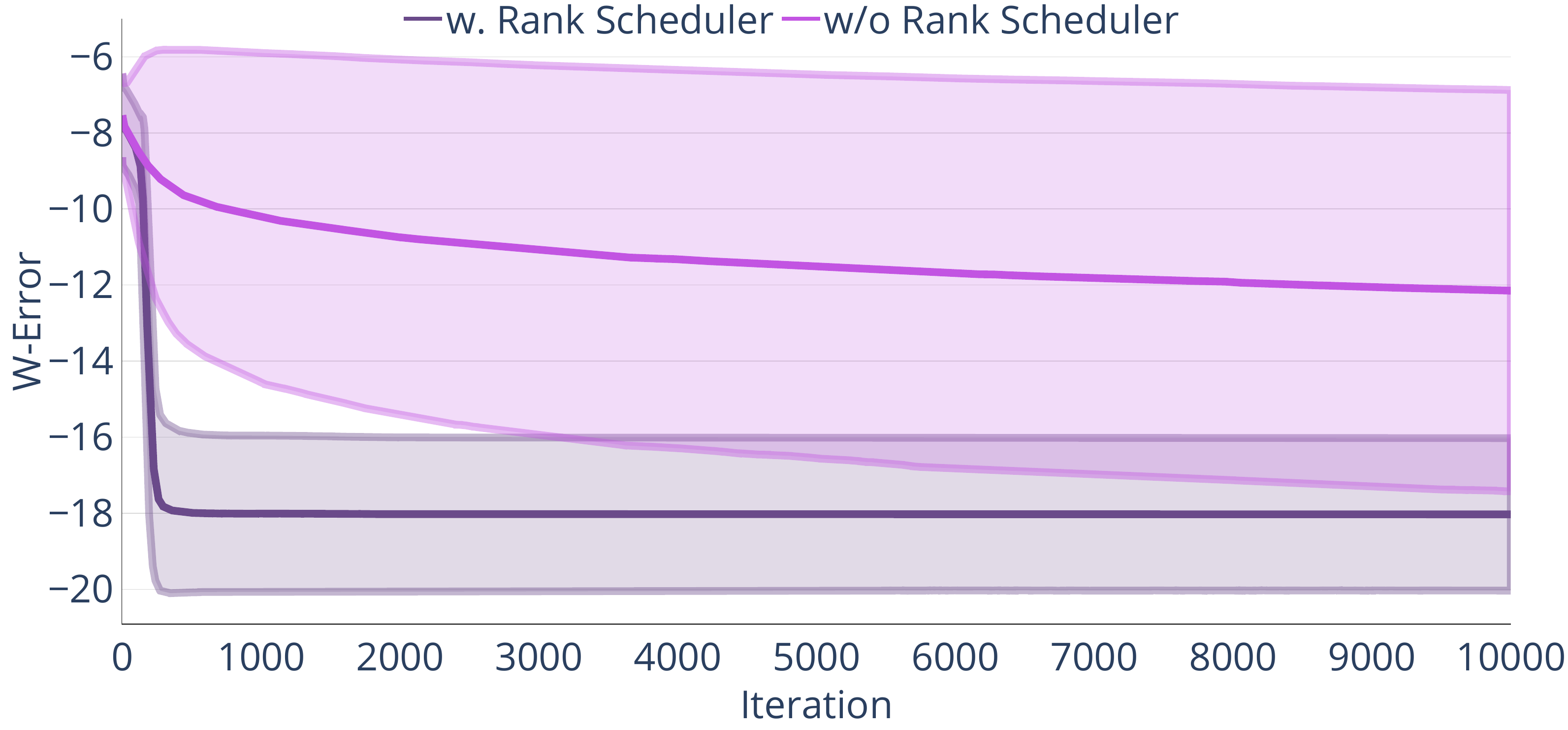}
 \caption{\textit{\textbf{Rank Scheduler Convergence Speed:}} Using the rank scheduler has three benefits, i) accelerated convergence , ii) less variance between layers, and iii) higher precision convergence. Here we visualize i), see \cref{fig:rank_sched_hist} for a layer-wise visualization}
\label{fig:rank_sched_vs_iters}
\end{figure}

\begin{figure}[t]
 \includegraphics[width=1.0\linewidth]{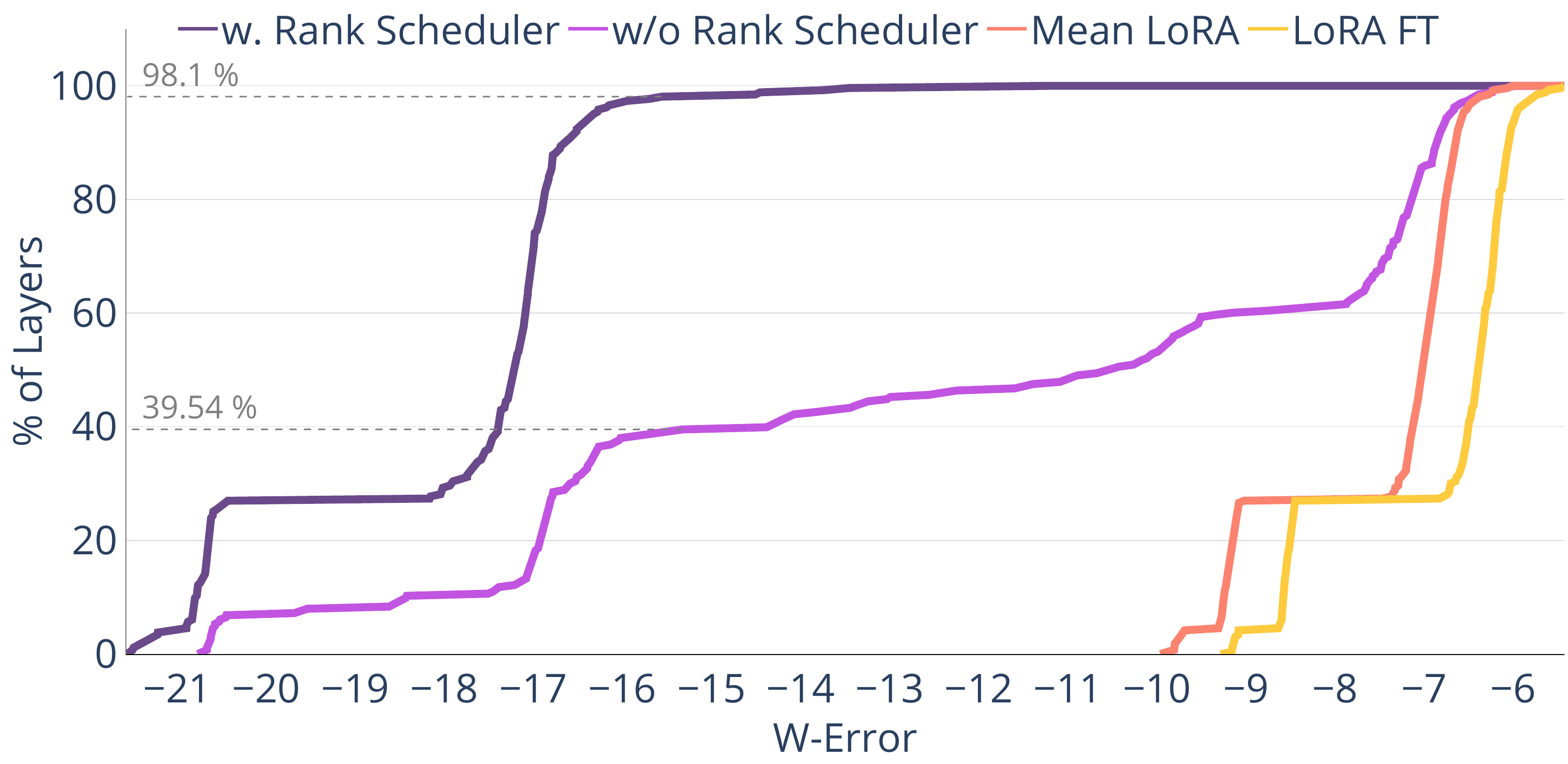}
 \caption{\textit{\textbf{Rank Scheduler Convergence Quality:}} When using the rank scheduler, over $95\%$ of the layers converge with a precision of at least $-16$, in contrast to less than $40\%$ without the scheduler}
\label{fig:rank_sched_hist}
\end{figure}

\noindent \textbf{Rank Scheduler Ablation.} We ablate the rank scheduler introduced in \cref{sec:rank_scheduler} using the Stable Diffusion experiment. Based on \cref{fig:rank_sched_vs_iters} we observe three phenomena, i) The rank scheduler drastically \textit{accelerates} the convergence, ii) When using the rank scheduler, there is much \textit{less variance} between the convergence of different layers, and iii) Using the rank scheduler results in a \textit{higher precision} convergence. \cref{fig:rank_sched_hist} visualizes phenomena (ii) and (iii) by showing the cumulative percent of layers (y axis) that converge to a given W-Error level (x axis). When using the rank scheduler, over $95\%$ of the layers converge with a precision of at least $-16$, in contrast to less than $40\%$ when not using the scheduler. Moreover, by using the rank scheduler, some layers converge to a more precise solution.

\noindent \textbf{Robustness to Unknown and Varying Ranks.} We tested the LoRA rank estimation heuristic presented in \cref{sec:rank_estimation} on hundreds of combinations of LoRAs with different ranks. The heuristic achieved an accuracy of $100\%$. We further tested the idea of using a dedicated rank scheduler for each LoRA model as described in \cref{sec:rank_scheduler,sec:rank_estimation}. We use $n=6$ fine-tuned LoRAs with ranks $[8,32,32,32,64,100]$ taken from CivitAI. Spectral DeTuning is robust to the varying ranks, exhibiting only a minor decrease in performance despite the higher rank of the LoRAs (See \cref{table:varying_ranks_ablation}).

\noindent \textbf{Robustness to Different Models.} We demonstrate the robustness of Spectral DeTuning to cases where a fine-tuned LoRA from a different Pre-FT model (with the same architecture) was mixed into the set of fine-tuned LoRAs. Using the same heuristic presented in \cref{sec:rank_estimation}, the difference between the mixed model weights and any other LoRA should be of full rank (since the Pre-FT model is different) and trivial to detect.

We validated this solution using Stable Diffusion. We added to the set of fine-tuned LoRA models a model that originated from Stable Diffusion 1.4 (all the others originated from Stable Diffusion 1.5). Indeed, the above steps indicated the LoRA that originated from Stable Diffusion 1.4 has a full rank difference from any other LoRA (while the pairwise rank between the LoRAs that used the same Pre-FT model were low rank, as expected). This allows us to detect the LoRA that got mixed up into the set and remove it.

\begin{table}[t]
    \caption{\textit{\textbf{Robustness to Unknown and Varying Ranks Results:}} We test the robustness to LoRAs with varying ranks. Spectral DeTuning is robust to varying ranks, exhibiting only a minor decrease in performance. We use $n=6$ fine-tuned LoRAs with ranks $[8,32,32,32,64,100]$ taken from an online LoRA marketplace}
    \begin{center}
    \begin{tabular}{l@{\hskip5pt}c@{\hskip5pt}c@{\hskip5pt}}    
            Method  & W-Error $\downarrow$   & LPIPS $\downarrow$\\
            \midrule
            LoRA FT  & -5.882 & 0.462 \\
            Mean LoRA  & -6.969 & 0.307 \\
            Spectral DeTuning  & \textbf{-14.453} & \textbf{0.073} \\
        \end{tabular}
        \end{center}
        \label{table:varying_ranks_ablation}
\end{table}

\noindent \textbf{W-Error vs. Loss.}  In reality an attacker has no access to the error and can only measure the loss in \cref{eq:spectral_detuning_optimization}. To show the loss accurately reflects the error defined in \cref{eq:pre_ft_objective}, we measure their relation and find they are almost perfectly correlated ($\rho=0.994$). For further details see \cref{app:w_error_vs_log_loss}.

\section{Discussion and Limitations}
\label{sec:discussion}
\begin{figure}[t]
 \includegraphics[width=1.0\linewidth]{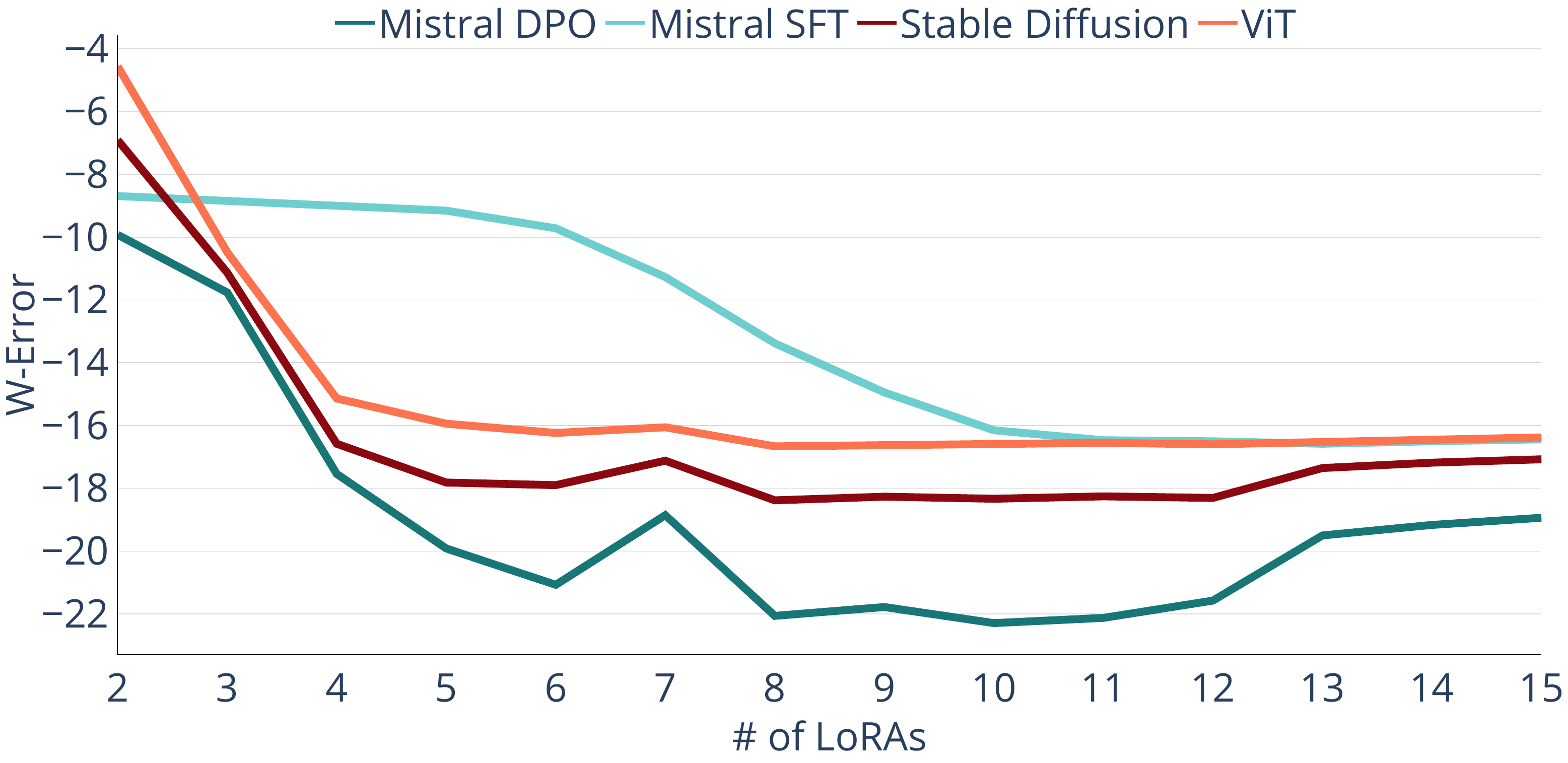}
 \caption{\textit{\textbf{Effect of the Number of LoRAs on W-Error Convergence:}} For the semantic equivalent see \cref{app:semantic_vs_n_loras}}
\label{fig:w_error_vs_n_loras}
\end{figure}

\noindent \textbf{Number of LoRAs.} Spectral DeTuning requires several LoRAs to recover the Pre-FT weights. In \cref{fig:w_error_vs_n_loras} we illustrate the impact of the number of fine-tuned LoRA models on the W-Error convergence. 
Note that different W-Error values are not comparable across models, e.g., Mistral DPO obtains a lowest W-Error but only semantically converges when using 8 LoRAs (See \cref{fig:n_loras_vs_semantic_dpo}). In \cref{app:semantic_vs_n_loras} we study the effects of the number of LoRAs on the semantic convergence for all LoWRA Bench subsets. We anticipate that future methods will incorporate additional constraints to reduce the required number of LoRAs.

\noindent \textbf{Public Availability of LoRA Fine-tuned Models.}
We assume the availability of multiple LoRA fine-tuned models originating from the same pre-fine-tuning model. This is a reasonable assumption as there are model \q{marketplaces} such as \textit{Hugging Face} and \textit{civitai}, where many LoRA fine-tuned models are publicly available. These LoRA models often share the same source Pre-FT model, which fits our proposed setting perfectly.

\noindent \textbf{Other Types of Fine-tuning.} While our focus has been on exposing the vulnerability of LoRA fine-tuned models, numerous other parameter-efficient fine-tuning methods exist. The general case of Pre-FT weight recovery of fully fine-tuned models is the most general and probably hardest case. Extending the scope of our attack to encompass these methods presents an exciting avenue for research.

\noindent \textbf{Pre-FT Weight Recovery Defense.} We do not know of a defense against this attack. Also, as this attack targets publicly available models, once a vulnerability is identified, there is no option to retract the model. However, we remain optimistic that a defense will be discovered in the future. For instance, modifying training such that an infeasible high number of LoRAs will be required for accurate recovery.

\section{Conclusion}
\label{sec:conclusion}
In this paper, we unveiled a new vulnerability in LoRA fine-tuned models, allowing attackers to recover the Pre-FT weights using multiple models. Our method, Spectral DeTuning, demonstrates this vulnerability on large-scale models like Mistral and Stable Diffusion. We introduced LoWRA Bench and discussed future directions to promote further research. By highlighting this vulnerability, we hope to encourage the research community to develop better defenses against such attacks.

\section{Acknowledgements}
This work was supported in part by the \q{Israel Science Foundation} (ISF), the \q{Council for Higher Education} (Vatat), and the \q{Center for Interdisciplinary Data Science Research} (CIDR).

\section{Broader Impact}
\label{sec:broader_impact}
This work uncovers a significant vulnerability in fine-tuned models, allowing attackers to access pre-fine-tuning weights. While this discovery reveals potential security risks, our primary objective is to advance the field of Machine Learning and raise awareness within the research community about the existing vulnerabilities in current models.

Instead of using the findings of this study to execute attacks, we advocate for their use by model creators to enhance the safety and security of their models. By acknowledging and addressing vulnerabilities, creators can proactively safeguard against potential threats.

Furthermore, in the discussion section, we outline potential future directions and mitigation strategies. Following established practices in the cyber security community, we emphasize the importance of open discussion and encourage the reporting of vulnerabilities. By fostering transparency and collaboration, we can collectively create a safer environment for deploying machine learning models.

\newpage

{
    \small
    \bibliographystyle{ieeenat_fullname}
    \bibliography{main}
}

\newpage
\appendix
\onecolumn

\section{The Effect of the Number of LoRAs on Semantic Convergence}
\label{app:semantic_vs_n_loras}
\begin{figure*}[h!]
    \begin{minipage}{.49\linewidth}
        \centering
        \includegraphics[width=1.0\linewidth]{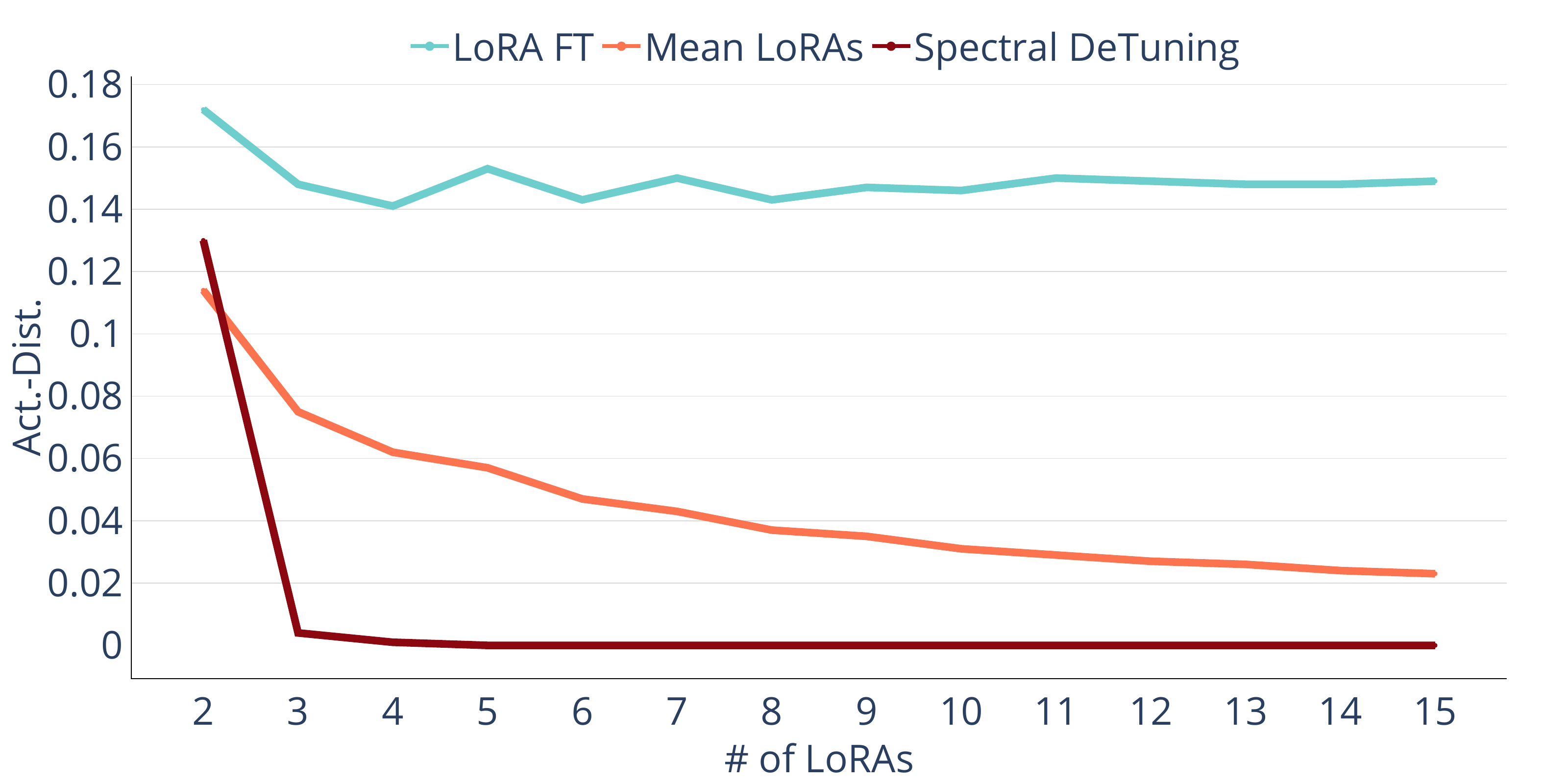}
        \caption{\textit{\textbf{Number of LoRAs vs. Semantic Convergence - ViT}}}
        \label{fig:n_loras_vs_semantic_vit}
    \end{minipage}%
    \hfill %
    \begin{minipage}{.49\linewidth}
        \centering
        \includegraphics[width=1.0\linewidth]{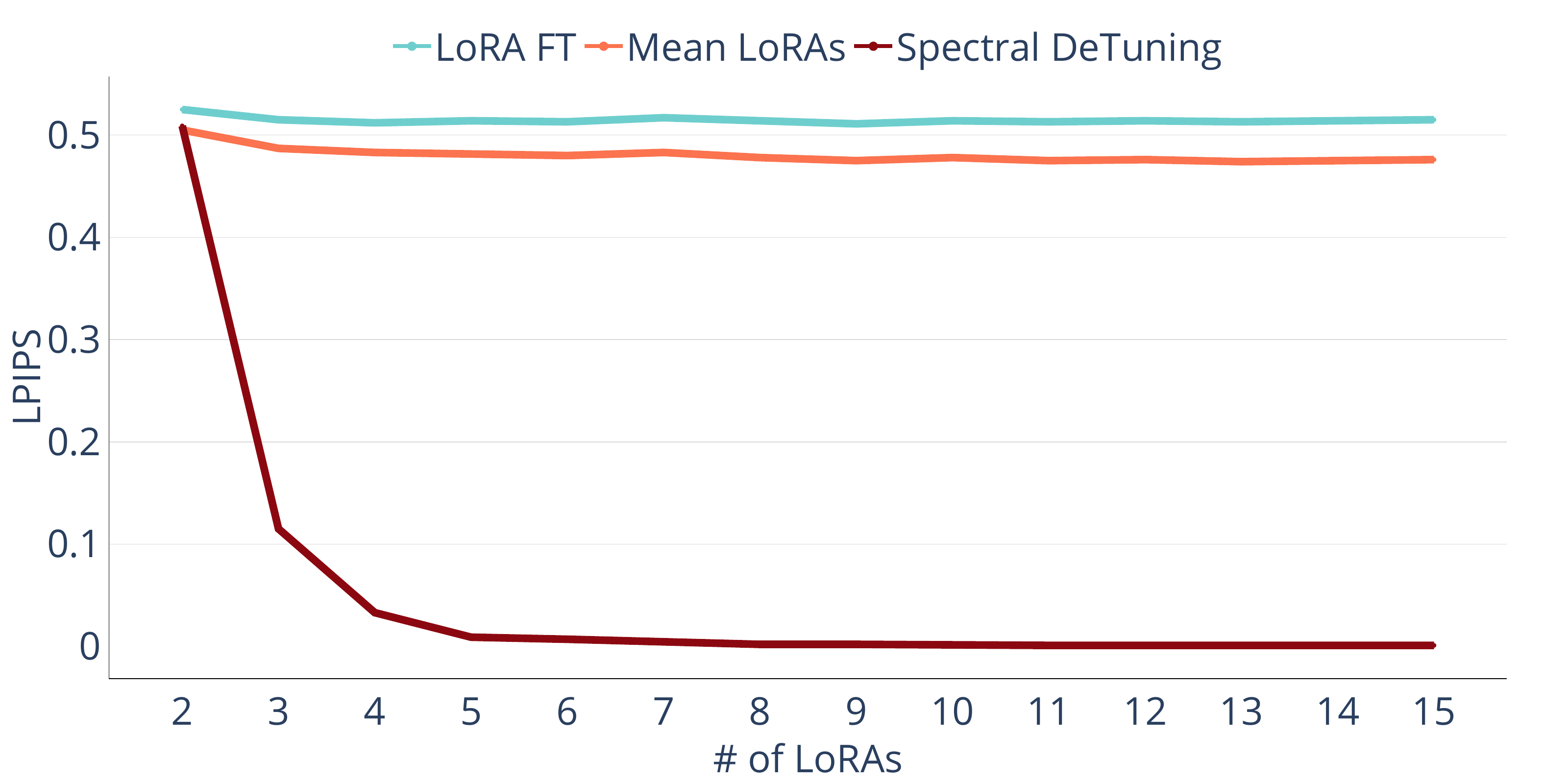}
        \caption{\textit{\textbf{Number of LoRAs vs. Semantic Convergence - Stable Diffusion}}}
        \label{fig:n_loras_vs_semantic_sd}
    \end{minipage} \\
    \begin{minipage}{.49\linewidth}
        \centering
        \includegraphics[width=1.0\linewidth]{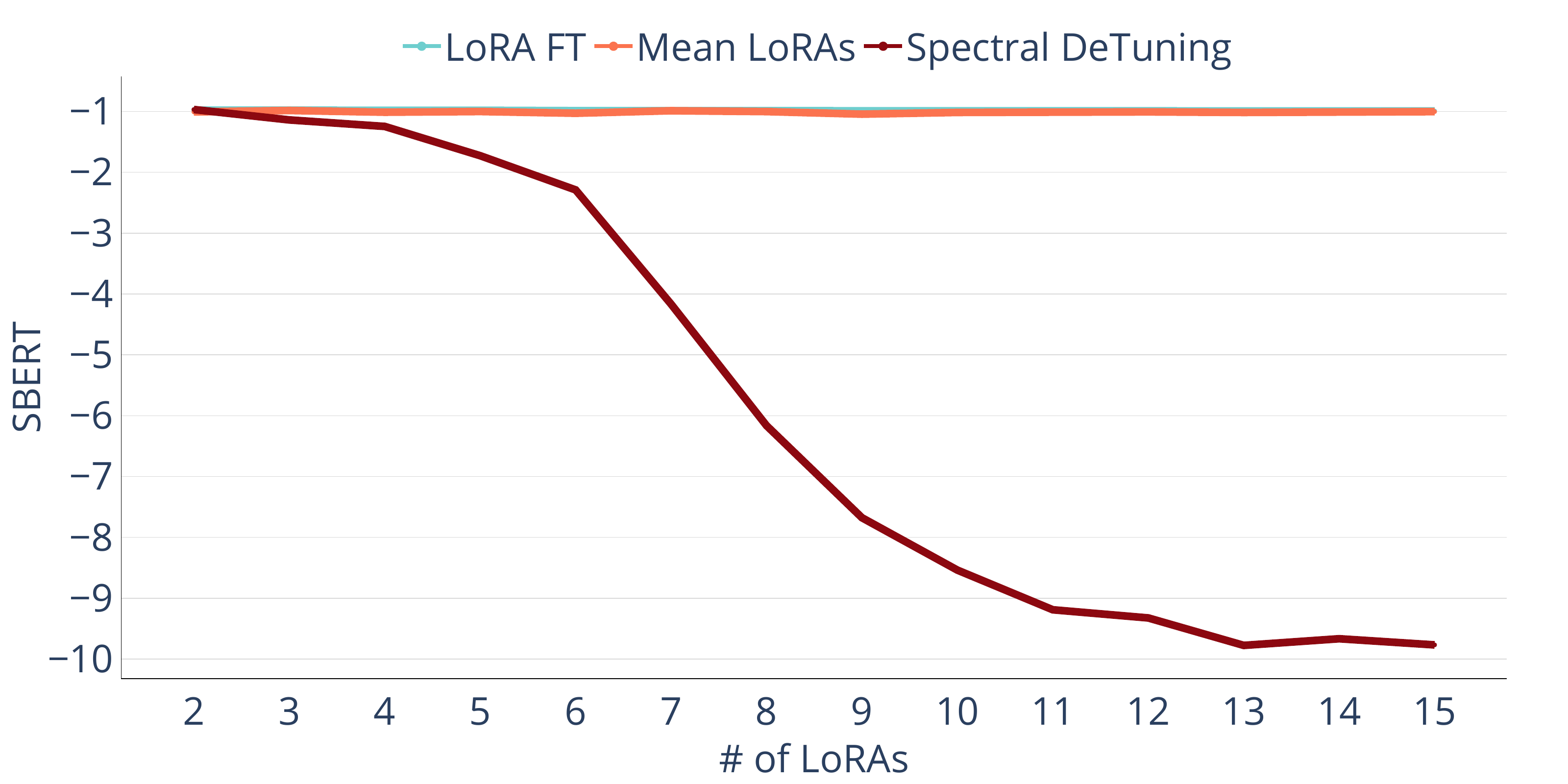}
        \caption{\textit{\textbf{Number of LoRAs vs. Semantic Convergence - Mistral SFT}}}
        \label{fig:n_loras_vs_semantic_sft}
    \end{minipage}%
    \hfill %
    \begin{minipage}{.49\linewidth}
        \centering
        \includegraphics[width=1.0\linewidth]{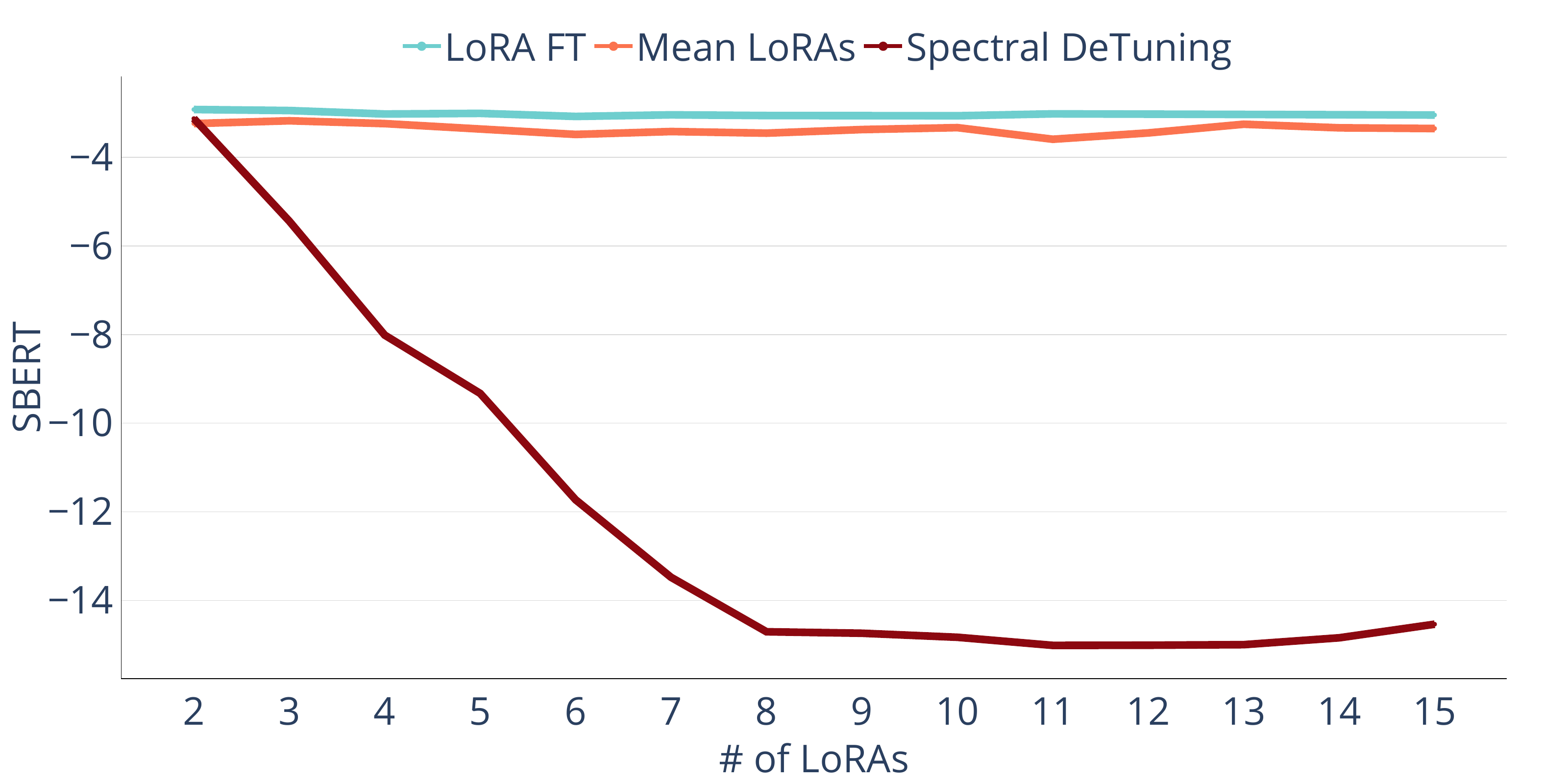}
        \caption{\textit{\textbf{Number of LoRAs vs. Semantic Convergence - Mistral DPO}}}
        \label{fig:n_loras_vs_semantic_dpo}
    \end{minipage} 
\end{figure*}

We visualize the effect of the number of LoRAs on the semantic convergence for each of the LoWRA Bench subsets, results are shown in \cref{fig:n_loras_vs_semantic_vit,fig:n_loras_vs_semantic_sd,fig:n_loras_vs_semantic_sft,fig:n_loras_vs_semantic_dpo}.

\section{W-Error vs. Loss}
\begin{figure*}[h!]
    \begin{minipage}{.49\linewidth}
        \centering
        \includegraphics[width=1.0\linewidth]{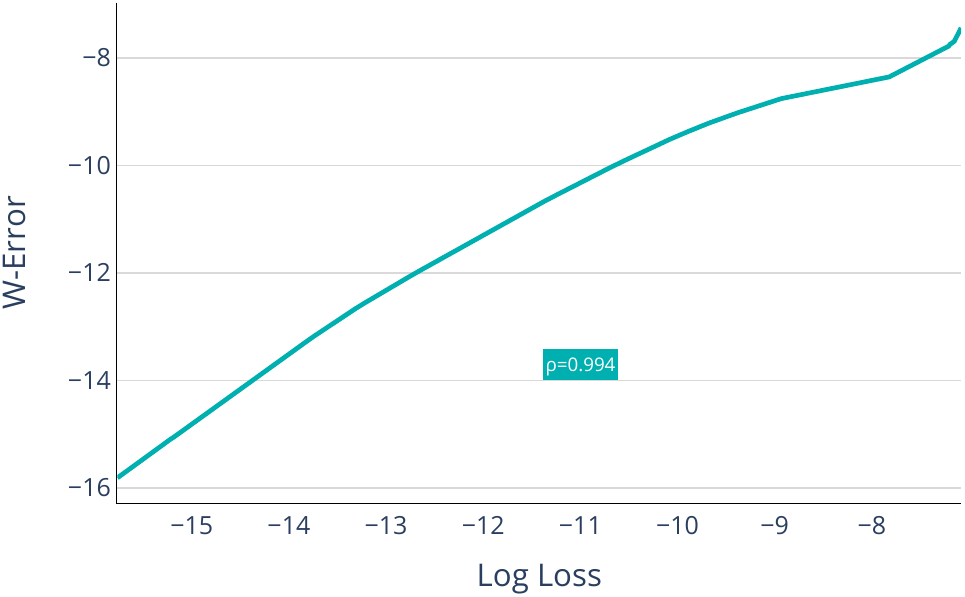}
        \caption{\textit{\textbf{W-Error vs. Loss - Stable Diffusion}}}
        \label{fig:w_error_vs_log_loss}
    \end{minipage}%
    \hfill %
    \begin{minipage}{.49\linewidth}
        \centering
        \includegraphics[width=1.0\linewidth]{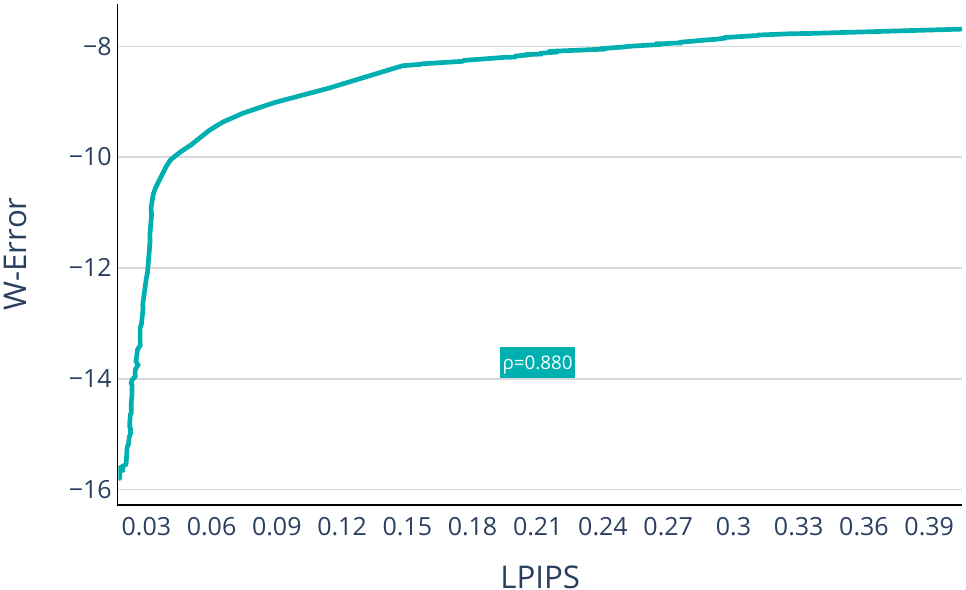}
        \caption{\textit{\textbf{W-Error vs. LPIPS}}}
        \label{fig:lpips_vs_w_error}
    \end{minipage} 
\end{figure*}

We visualize the relation between the W-Error and the log loss and find they are almost perfectly correlated ($\rho=0.994$), see \cref{fig:w_error_vs_log_loss} for a visualization over $200$ iterations using Stable Diffusion. 
\label{app:w_error_vs_log_loss}

\section{W-Error vs. LPIPS}
\label{app:w_error_vs_lpips}
We visualize the relation between the W-Error and LPIPS and find they are strongly correlated ($\rho=0.880$), see \cref{fig:lpips_vs_w_error} for a visualization over $200$ iterations using Stable Diffusion.

\section{LoRA Rank vs. W-Error} In \cref{table:ranks_vs_werror_5,table:ranks_vs_werror_6} we show the results for the ViT model when using different LoRA ranks and fixing the number of LoRAs.

\begin{table*}[h!]
    \begin{minipage}{.49\linewidth}
      \caption{\textit{\textbf{Using $5$ LoRAs}}}
        \centering
    \begin{tabular}{c@{\hskip5pt}l@{\hskip5pt}}    
         Rank & W-Error \\
         \midrule
         8 & $-15.636$ \\
         12 & $-15.550$ \\
         16 & $-13.480$ \\
         32 & $-4.817$ \\
    \end{tabular}
    \label{table:ranks_vs_werror_5}
    \end{minipage}%
    \hfill %
    \begin{minipage}{.49\linewidth}
       \caption{\textit{\textbf{Using $5$ LoRAs}}}
        \centering
 \begin{tabular}{c@{\hskip5pt}l@{\hskip5pt}}    
         Rank & W-Error \\
         \midrule
         8 & $-15.822$ \\
         12 & $-15.773$ \\
         16 & $-15.258$ \\
         32 & $-9.639$ \\
    \end{tabular}
        \label{table:ranks_vs_werror_6}
    \end{minipage} 
\end{table*}

\section{LoWRA Bench Dataset}
\label{app:lowra_bench}
We now elaborate on the implementation details of the LoWRA Bench dataset.

\subsection{ViT Models}
\label{app:lowra_bench_vit}
As the Pre-FT model we use \q{vit-base-patch16-224} found on hugging face (\url{https://huggingface.co/google/vit-base-patch16-224}). We fine-tune the model using the PEFT library \cite{hugging_face_peft}. For each LoRA we use a different VTAB-1k \cite{vtab} dataset, the datasets we use are: \texttt{cifar100, caltech101, dtd, flower102, pet37, svhn, patch\_camelyon, clevr-count, clevr-distance, dmlab, kitti, dsprites-location, dsprites-orientation, smallnorb-azimuth, smallnorb-elevation}. We pre-process the datasets according to the protocol of \citet{vpt} found on their github page \url{https://github.com/KMnP/vpt/blob/main/VTAB_SETUP.md}. We use an $80/20$ train/validation split and choose the checkpoint with the best validation loss.

We use a rank $r=16$ and LoRA fine-tune the \texttt{query} and \texttt{value} layers. This protocol results in $24$ Pre-FT model layers and a total of $24\cdot15=360$ LoRA fine-tuned layers. See \cref{table:vit_hyperparams} for the fine-tuning hyper-parameters. 

For semantic evaluation we use a subset of the ImageNet-1K \cite{imagenet} validation set. We construct the subset by taking the first $5$ images of each class, resulting in a subset of $5000$. 

\begin{table}[h!]
\caption{\textit{\textbf{ViT Hyper-parameters}}}
\begin{center}
    \begin{tabular}{l@{\hskip5pt}c@{\hskip5pt}}    
         Name & Value \\
         \midrule
         \texttt{lora\_rank} ($r$) & $16$ \\
         \texttt{lora\_alpha} ($\alpha$) & $16$ \\
         \texttt{lr} & $9e-3$ \\
         \texttt{batch\_size} & $128$ \\
         \texttt{epochs} & $20$ \\
         \texttt{datasets} & \begin{tabular}[c]{@{}c@{}}cifar100, caltech101, dtd, flower102, pet37, svhn, patch\_camelyon,\\ clevr-count, clevr-distance, dmlab, kitti, dsprites-location, dsprites-orientation,\\ smallnorb-azimuth, smallnorb-elevation \end{tabular}\\
    \end{tabular}
\end{center}
\label{table:vit_hyperparams}
\end{table}

\subsection{Mistral Models}
\label{app:lowra_bench_mistral}
As the Pre-FT model we use \q{Mistral-7B-v0.1} found on hugging face (\url{https://huggingface.co/mistralai/Mistral-7B-v0.1}). We fine-tune the model following the protocol of \citet{zephyr}, note that unlike \citet{zephyr}, we perform LoRA fine-tuning as found on their official github repo \url{https://github.com/huggingface/alignment-handbook}. Following the original LoRA setting, we make a minor adjustment to the original hyper-parameters of the repo and use a LoRA alpha of $64$ instead of $16$ (i.e. $\alpha=64$), this leads to faster and better convergence. To fine-tune $15$ different models, we use different \textit{random} subsets of $80\%$ of the fine-tuning dataset. We use seeds of $0-14$ for the different fine-tuned models. 

We follow this protocol for both the supervised fine-tuning stage (SFT) and the direct preference optimization (DPO) alignment stage. Following \citet{zephyr}, the SFT stage uses the UltraChat dataset \citep{ultrachat_dataset} and the DPO stage uses the UltraFeedback dataset \citep{ultrafeedback_dataset}. We first fine-tune the $15$ SFT models, and then fine-tune the $15$ DPO models, where each DPO model continues the training of the SFT model with the corresponding seed.

Following the original setup, use a rank $r=64$ and LoRA fine-tune the \texttt{q\_proj}, \texttt{k\_proj}, \texttt{v\_proj}, and \texttt{o\_proj} layers. This protocol results in $128$ Pre-FT model layers and a total of $128\cdot15=1920$ LoRA fine-tuned layers for both the SFT and DPO stages. See \cref{table:mistral_sft_hyperparams,table:mistral_dpo_hyperparams} for the fine-tuning hyper-parameters. For inference we use the following decoding hyper-parameters: \texttt{max\_new\_tokens=50, do\_sample=True, temperature=0.7, top\_k=50, top\_p=0.95}.

For evaluation we use the first $100$ prompts from the AlpacaFarm benchmark \cite{alpaca_farm} found in the following link \url{https://huggingface.co/datasets/tatsu-lab/alpaca_farm/viewer/alpaca_farm_evaluation}. We provide these prompts in the SM.

\begin{table*}[h!]
    \begin{minipage}{.49\linewidth}
      \caption{\textit{\textbf{Mistral SFT Hyper-parameters}}}
        \centering
    \begin{tabular}{l@{\hskip5pt}l@{\hskip5pt}}    
         Name & Value \\
         \midrule
         \texttt{lora\_rank} ($r$) & $64$ \\
         \texttt{lora\_alpha} ($\alpha$) & $64$ \\
         \texttt{lora\_dropout} & $0.1$ \\
         \texttt{lr} & $2e-5$ \\
         \texttt{batch\_size} & $4$ \\
         \texttt{gradient\_accumulation\_steps} & $128$ \\
         \texttt{learning\_rate\_scheduler} & Cosine \\
         \texttt{epochs} & $1$ \\
         \texttt{warmup\_ratio} & $0.1$ \\
         \texttt{data\_type} & bfloat16 \\
         \texttt{dataset} & \begin{tabular}[l]{@{}l@{}}random $80\%$ of\\ UltraChat \end{tabular}\\
         \texttt{seeds} & $0-15$ \\
    \end{tabular}
    \label{table:mistral_sft_hyperparams}
    \end{minipage}%
    \hfill %
    \begin{minipage}{.49\linewidth}
       \caption{\textit{\textbf{Mistral DPO Hyper-parameters}}}
        \centering
 \begin{tabular}{l@{\hskip5pt}l@{\hskip5pt}}    
         Name & Value \\
         \midrule
         \texttt{lora\_rank} ($r$) & $64$ \\
         \texttt{lora\_alpha} ($\alpha$) & $64$ \\
         \texttt{lora\_dropout} & $0.1$ \\
         \texttt{lr} & $5e-6$ \\
         \texttt{batch\_size} & $2$ \\
         \texttt{gradient\_accumulation\_steps} & $32$ \\
         \texttt{learning\_rate\_scheduler} & Cosine \\
         \texttt{epochs} & $1$ \\
         \texttt{warmup\_ratio} & $0.1$ \\
         \texttt{data\_type} & bfloat16 \\
         \texttt{dataset} & \begin{tabular}[c]{@{}c@{}}random $80\%$ of\\ UltraFeedback \end{tabular}\\
         \texttt{seeds} & $0-15$ \\
    \end{tabular}
        \label{table:mistral_dpo_hyperparams}
    \end{minipage} 
\end{table*}

\subsection{Stable Diffusion Models}
\label{app:lowra_bench_stable_diffusion}
As the Pre-FT model we use \q{Stable Diffusion 1.5} found on hugging face (\url{https://huggingface.co/runwayml/stable-diffusion-v1-5}). We collect $15$ personalization fine-tuned models from civitai.com, a public and widely used LoRA models marketplace. This allows us to examine our method in a real world setting, for the full list of LoRAs see \cref{table:sd_lora_links}. After examining the downloaded models, we deduce that their LoRA rank is $r=32$ and that their fine-tuned layers are: \texttt{to\_q}, \texttt{to\_v}, \texttt{to\_k}, \texttt{to\_out}, \texttt{proj\_out}, \texttt{proj\_in}, and \texttt{ff}. Resulting in $192$ Pre-FT model layers for and a total of $192\cdot15=2880$ LoRA fine-tuned layers.  For inference we use the default Stable Diffusion $1.5$ generation pipeline (i.e. $50$ sampling steps).

For evaluation we use a the first $100$ captions from the COCO Captions \cite{coco_captions} validation dataset found in the following link \url{https://github.com/tylin/coco-caption/blob/master/annotations/captions_val2014.json}. We provide these prompts in the SM.

\begin{table}[h!]
\caption{\textit{\textbf{Stable Diffusion Fine-tuned LoRA Links}}}
\begin{center}
    \begin{tabular}{l@{\hskip5pt}}    
         \url{https://civitai.com/models/186716/smol-animals-lora-15sdxl?modelVersionId=241137}\\
         \url{https://civitai.com/models/189905/pastry-lora-15sdxl?modelVersionId=241955}\\
         \url{https://civitai.com/models/191203/bastet-egypt-cat-style-lora-15sdxl?modelVersionId=243232}\\
         \url{https://civitai.com/models/190176/fur-pirates-lora-15sdxl?modelVersionId=241976}\\
         \url{https://civitai.com/models/211973/cigarette-style-lora-15sdxl?modelVersionId=247079}\\
         \url{https://civitai.com/models/233316/smol-dragons-lora-15sdxl?modelVersionId=263316}\\
         \url{https://civitai.com/models/234324/polygon-style-lora-15sdxl?modelVersionId=264506}\\
         \url{https://civitai.com/models/202128/overgrowth-style-lora-15sdxl?modelVersionId=264449}\\
         \url{https://civitai.com/models/218327/mythical-creatures-lora-15sdxl?modelVersionId=289861}\\
         \url{https://civitai.com/models/203169/lava-style-lora-15sdxl?modelVersionId=265372}\\
         \url{https://civitai.com/models/197998/chocolate-coffee-style-lora-15sdxl?modelVersionId=259150}\\
         \url{https://civitai.com/models/180780/crystals-lora-15sdxl?modelVersionId=238435}\\
         \url{https://civitai.com/models/196040/transparent-glass-body-lora-15sdxl?modelVersionId=245630}\\
         \url{https://civitai.com/models/199968/liquid-flow-style-lora-15sdxl?modelVersionId=259228}\\
         \url{https://civitai.com/models/206783/christmas-critters-lora-15sdxl?modelVersionId=275204}\\
    \end{tabular}
\end{center}
\label{table:sd_lora_links}
\end{table}

\section{Spectral DeTuning Implementation Details}
\label{app:impl_details}
For all semantic evaluations we use a seed of $0$ for all baselines and for our results. For both the ViTs and Stable Diffusion (SD) experiments we run Spectral DeTuning for $300$ optimization steps. For the Mistral SFT and DPO experiments we use $1000$ optimization steps. We base our rank scheduler implementation on the official PyTorch implementation of a the \texttt{ReduceLROnPlateau} learning rate scheduler \footnote{\url{https://pytorch.org/docs/stable/generated/torch.optim.lr_scheduler.ReduceLROnPlateau.html}}. We expand on the hyper-parameters of the rank scheduler in \cref{table:rank_scheduler_hyperparams}.

\begin{table}[h!]
\caption{\textit{\textbf{Spectral DeTuning Rank Scheduler Hyper-parameters}}}
\begin{center}
    \begin{tabular}{l@{\hskip5pt}c@{\hskip5pt}l@{\hskip5pt}}    
         Name & Value Used & Explanation\\
         \midrule
         \texttt{total\_steps} & $200$ for ViT and SD, $1000$ for Mistral & The total number of optimization steps \\
         \texttt{start\_rank} & $1$ & The rank to start the optimization from (i.e. $r^{*}$) \\
         \texttt{end\_rank} & $16$ for ViTs, $32$ for SD, $64$ for Mistral & \begin{tabular}[l]{@{}l@{}} The final rank of the scheduler \\ (i.e. $r$, the actual rank of the LoRA models) \end{tabular}\\
         \texttt{factor} & $2$ & The multiplicative factor to increase the rank by \\
         \texttt{patience} & $15$ & \begin{tabular}[l]{@{}l@{}}Number of scheduler steps with no improvement after\\ which rank will be increased. \end{tabular} \\
         \texttt{force\_end\_rank\_percent} & $0.5$ & \begin{tabular}[l]{@{}l@{}} Percent of the \texttt{total\_steps} after \\which \texttt{end\_rank} will be forced \end{tabular} \\
    \end{tabular}
\end{center}
\label{table:rank_scheduler_hyperparams}
\end{table}

\section{Runtime and Compute}
\label{app:log_error_ablation}
Since Spectral DeTuning does not pass any gradients through the model, it is highly parallelizable and can recover the weights of even large models (e.g., Mistral 7B) in minutes using a cluster of desktop-grade GPUs or even \textit{CPUs}. For example, using a cluster of  RTX2080 it can recover Mistral-7B in under five minutes.

\section{Detecting the Fine-Tuned Layers}
\label{app:detecting_finetuned_layers}
We note that it is easy to detect which layers were fine-tuned. This can simply be done by comparing the layers weights of $n$ different fine-tuned versions. The layers which have not been fine-tuned will be equal across all $n$ models, while the fine-tuned layers will have some variation between them.

\clearpage

\section{Algorithm with Rank Scheduler}
\label{app:alg_with_scheduler}
In \cref{app:spectral_detuning_with_scheduler} we present pytorch-like pseudocode for Spectral DeTuning that includes that rank scheduler.

\begin{algorithm}[h!]
   \caption{PyTorch Pseudocode for Spectral DeTuning}
   
   \label{app:spectral_detuning_with_scheduler}
   
    \definecolor{codeblue}{rgb}{0.25,0.5,0.5}
    \definecolor{codekw}{rgb}{0.85, 0.18, 0.50}
    \lstset{
  backgroundcolor=\color{white},
  basicstyle=\fontsize{7.1pt}{7.1pt}\ttfamily\selectfont,
  columns=fullflexible,
  breaklines=true,
  captionpos=b,
  commentstyle=\fontsize{7.1pt}{7.1pt}\color{codeblue},
  keywordstyle=\fontsize{7.1pt}{7.1pt}\color{codekw},
}
\begin{lstlisting}[language=python]
# W_ps: List of n fine-tuned weight matrices
# steps: Number of optimization steps
# r: LoRA rank

# Initialize rank scheduler
current_lora_rank = 1
rank_scheduler = LoRARankScheduler(start_rank=current_lora_rank, end_rank=r)

# Initialize W_star
W_s = torch.mean(torch.stack(W_ps), axis=0)

# Perform optimization
for step in range(steps):
    # M-step
    # Approximate each M^*_i (Eq. 5)
    M_s = [W_p - W_s for W_p in W_ps]

    # Truncate each M^*_i to rank <= r (Eq. 5)
    for i in range(len(M_s)):
        (U, S, V) = torch.svd_lowrank(M_s[i], q=current_lora_rank)
        M_s[i] = (U @ torch.diag_embed(S)) @  V.T

    
    # W-step
    # Approximate W_star (Eq. 7)
    W_s = [W_p - M_si for (W_p, M_si) in zip(W_ps, M_s)]
    W_s = torch.mean(torch.stack(W_s), axis=0)

    # Compute the current loss
    iteration_losses = [torch.mean((W_ps[i] - (W_s + M_s[i])) ** 2) for i in range(len(M_s))]
    loss = torch.mean(torch.stack(iteration_losses), axis=0)

    # Step the rank scheduler
    rank_scheduler.step(loss)
    current_lora_rank = rank_scheduler.current_rank
\end{lstlisting}
\end{algorithm}

\section{Mistral Additional Results}
\label{app:mistral_all_results}
For the list of mistral prompts see supplementary material (SM). In \cref{app:mistral_results} we show side-by-side results for $10$ randomly (\texttt{random\_seed=42}) sampled prompts from our evaluation dataset, using the Pre-FT recovered weights of the DPO fine-tuned Mistral model. See SM for the rest of the DPO results and for the SFT results.

\section{Stable Diffusion Additional Results}
\label{app:sd_all_results}
\begin{figure*}[th!]
    \includegraphics[width=1.0\linewidth]{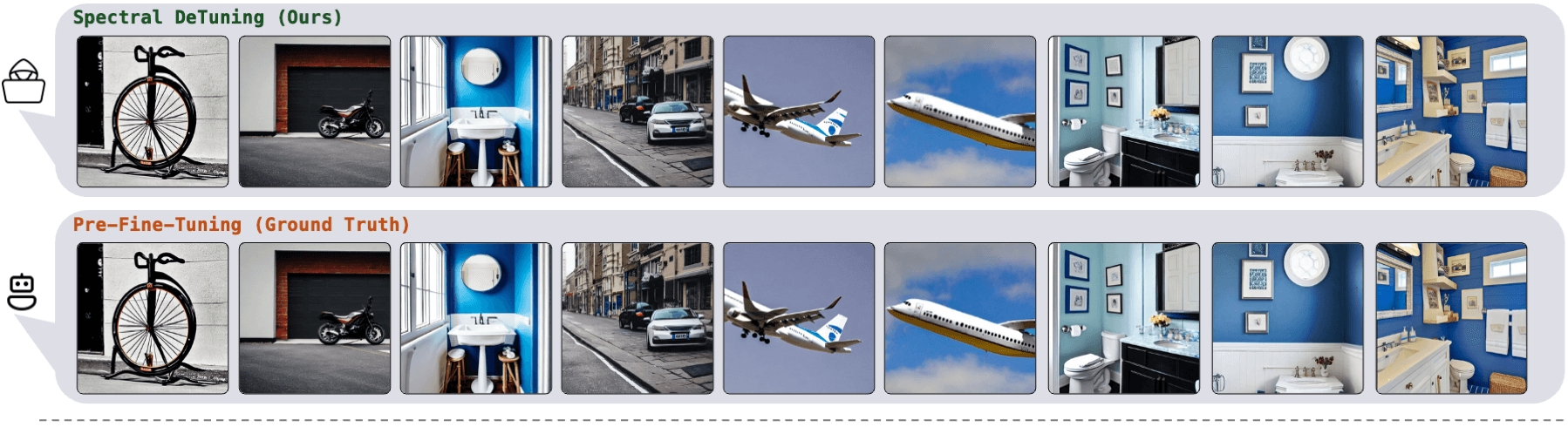} 
    \includegraphics[width=1.0\linewidth]{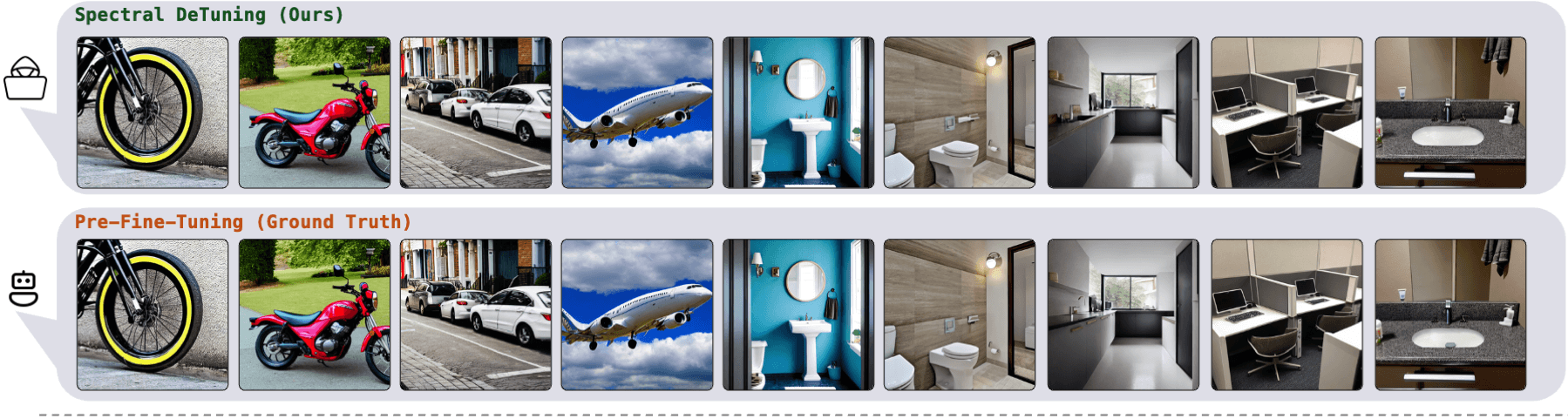}
    \includegraphics[width=1.0\linewidth]{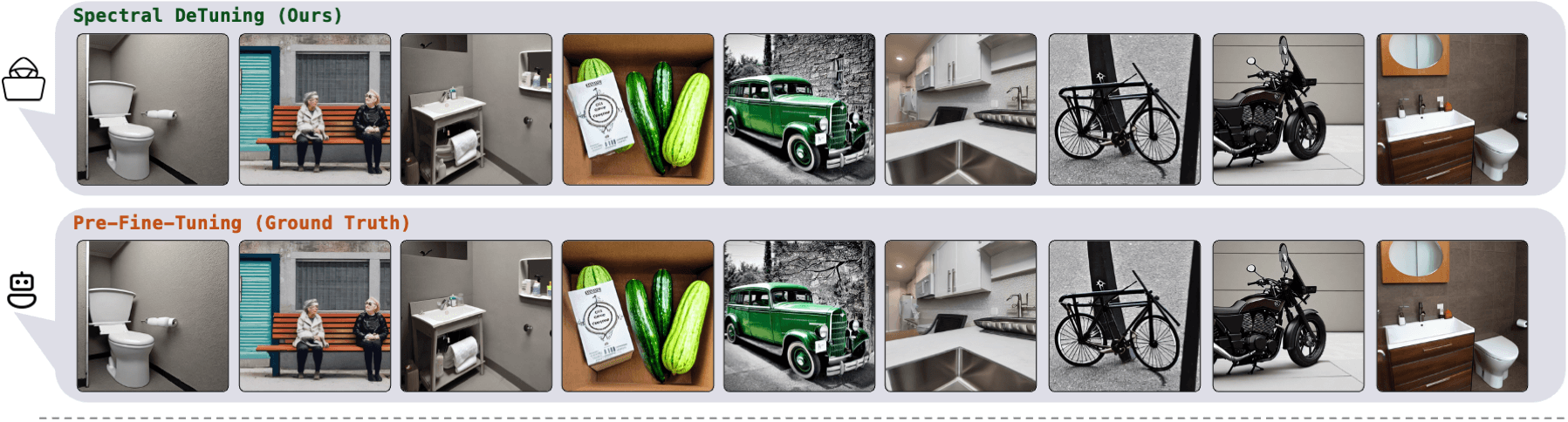}
    \includegraphics[width=1.0\linewidth]{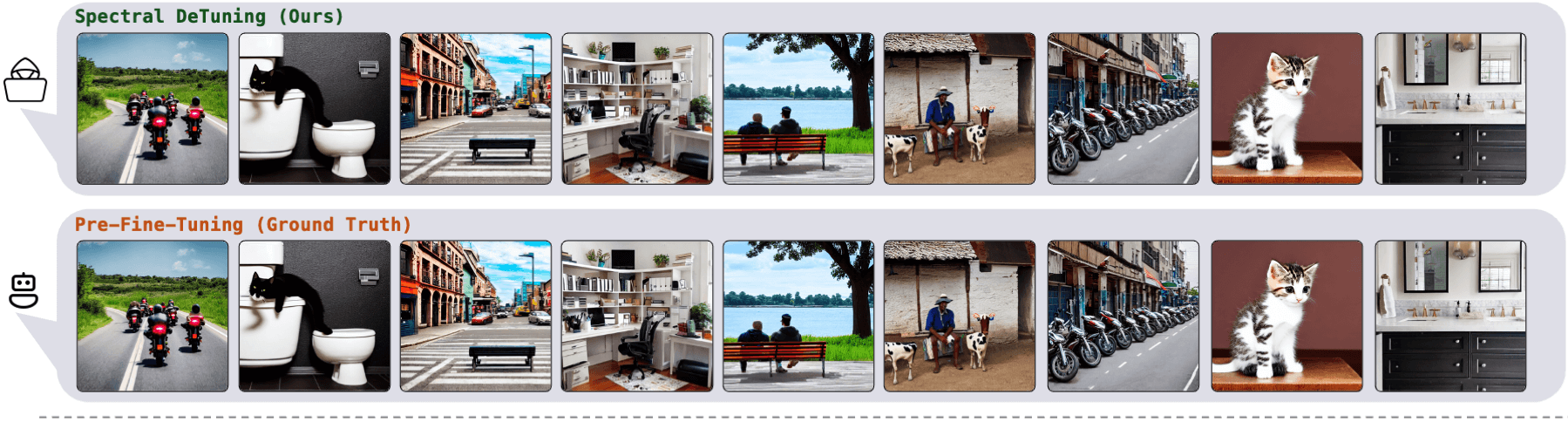}    
    \caption{\textit{\textbf{Stable Diffusion Results:}} Note, images are compressed to reduce file size, for the full resolution images see the SM.}
    \label{app:stable_diffusion_results1}
\end{figure*}

\begin{figure*}[th!]
    \includegraphics[width=1.0\linewidth]{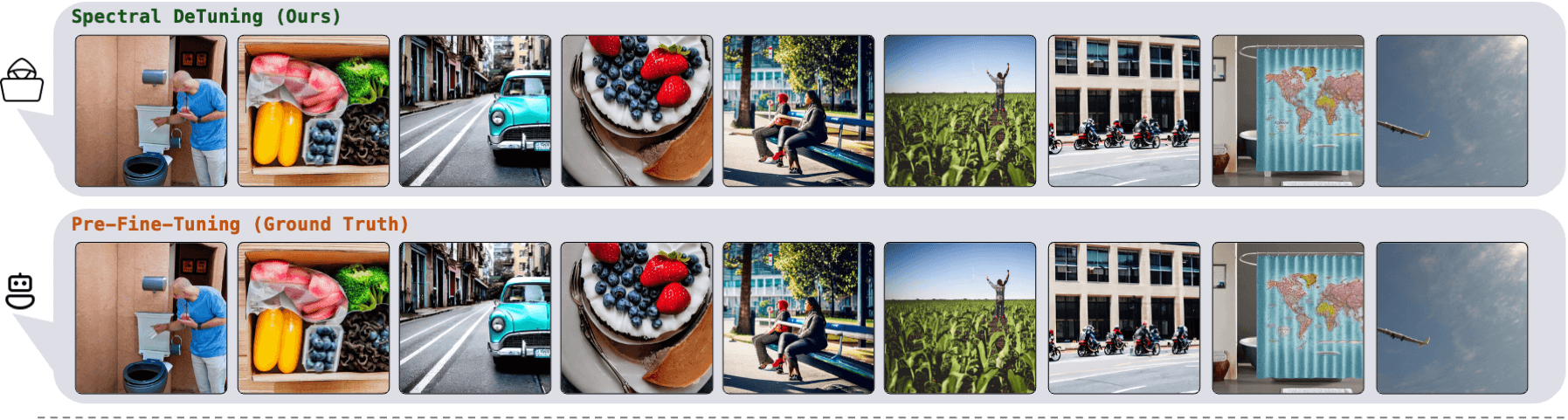}
    \includegraphics[width=1.0\linewidth]{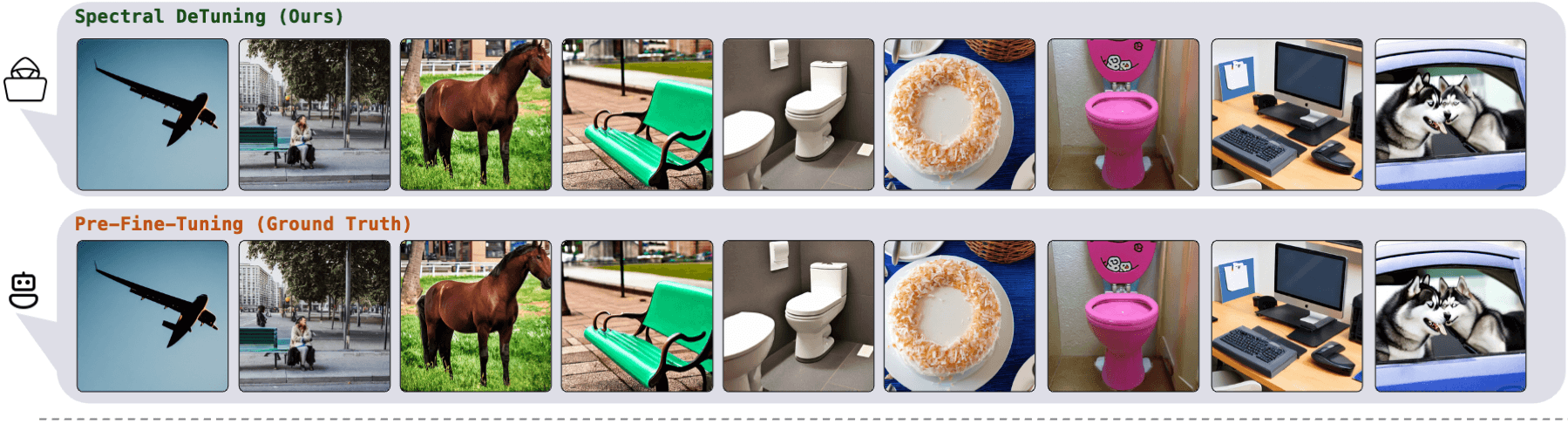}
    \includegraphics[width=1.0\linewidth]{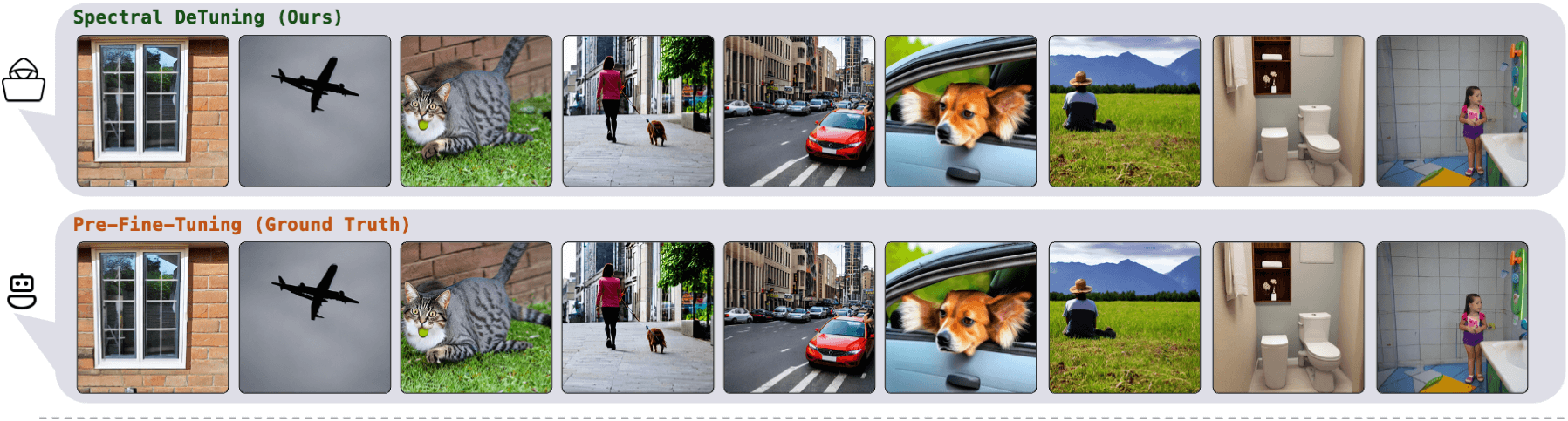}
    \includegraphics[width=1.0\linewidth]{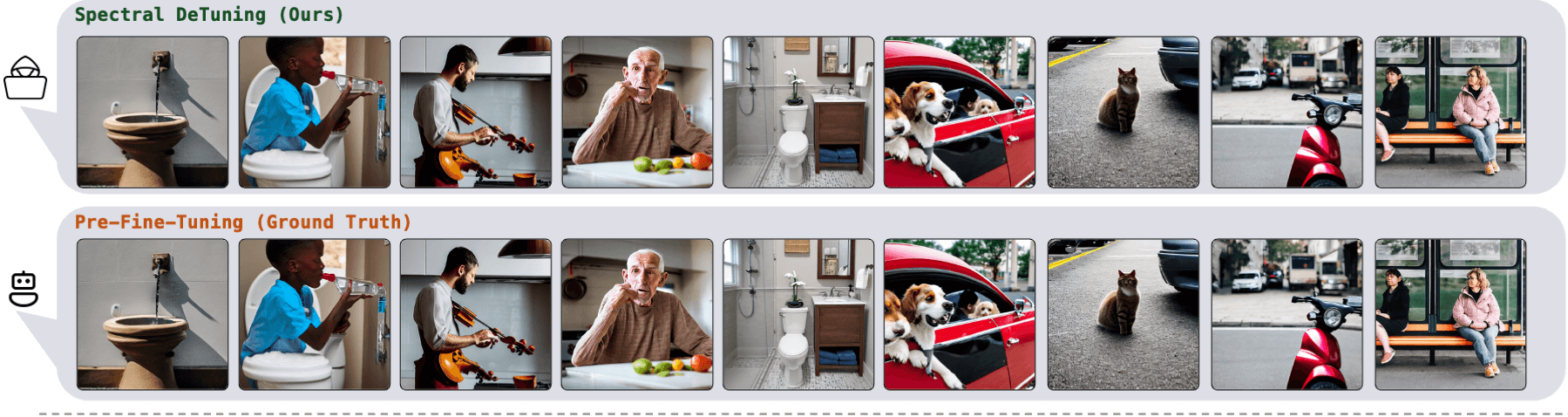}    
    \caption{\textit{\textbf{Stable Diffusion Results:}} Note, images are compressed to reduce file size, for the full resolution images see the SM.}
    \label{app:stable_diffusion_results2}
\end{figure*}

\begin{figure*}[th!]
    \includegraphics[width=1.0\linewidth]{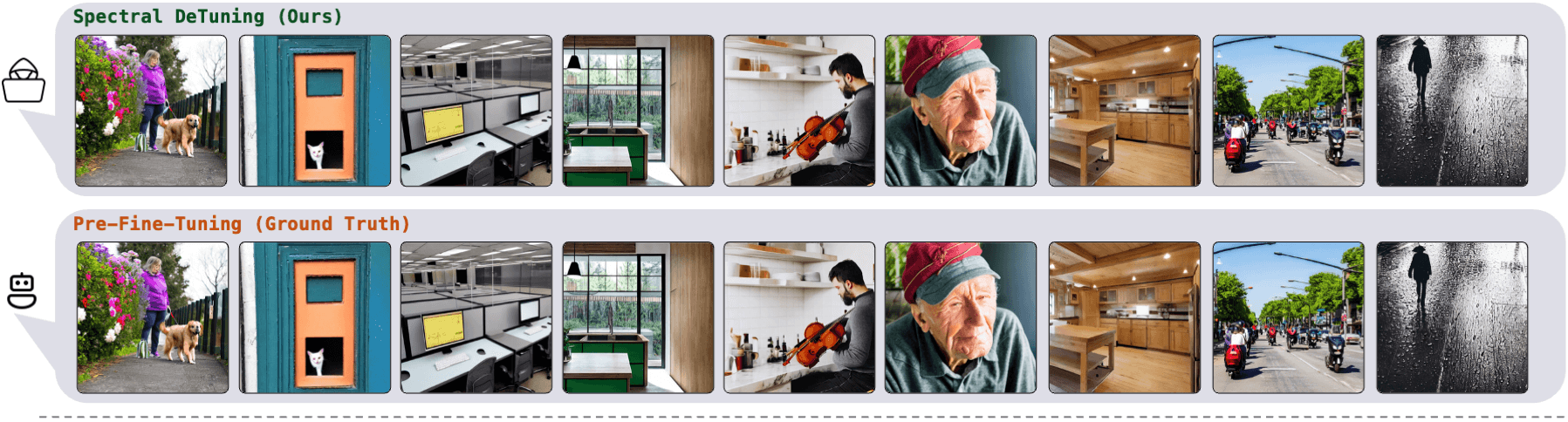}
    \includegraphics[width=1.0\linewidth]{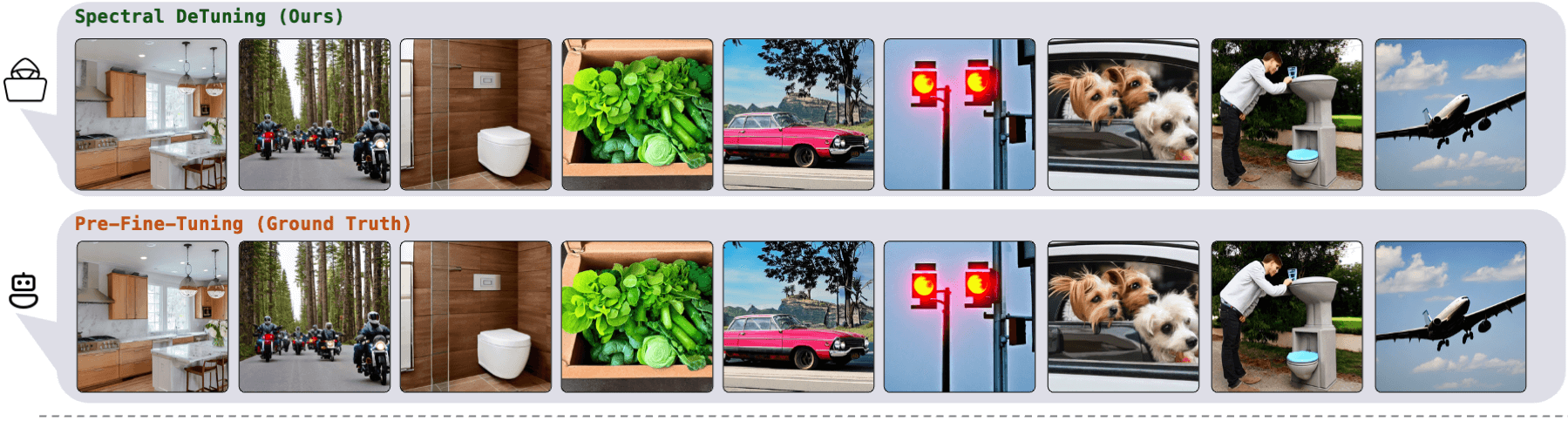}
    \includegraphics[width=1.0\linewidth]{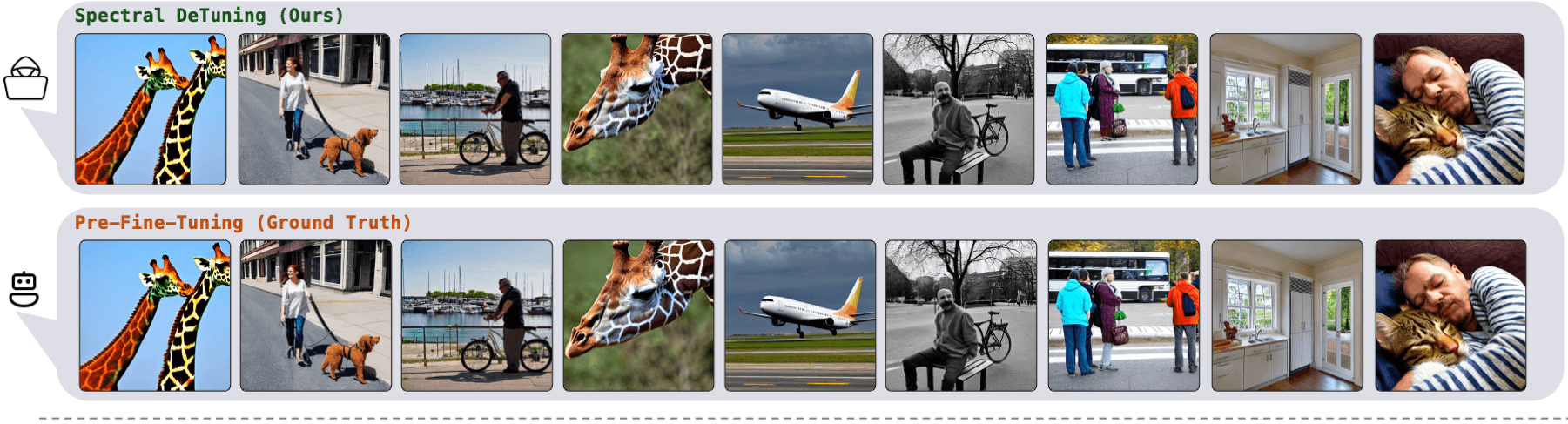}
    \includegraphics[width=1.0\linewidth]{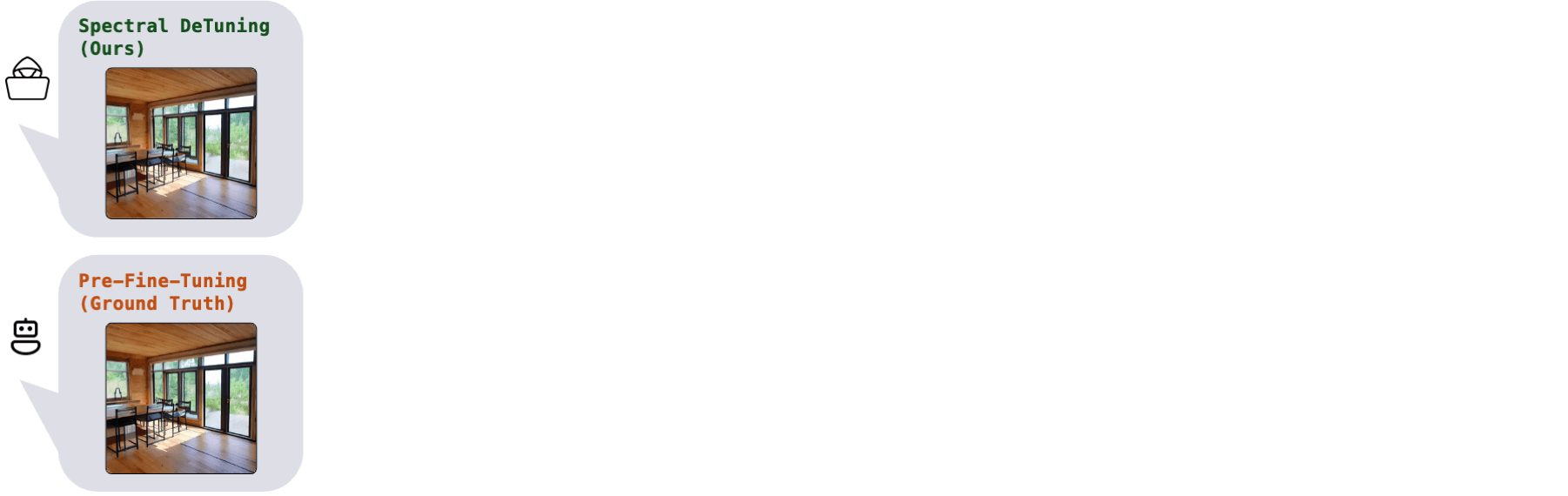}    
    \caption{\textit{\textbf{Stable Diffusion Results:}} Note, images are compressed to reduce file size, for the full resolution images see the SM.}
    \label{app:stable_diffusion_results3}
\end{figure*}

For the list of stable diffusion prompts see SM. In \cref{app:stable_diffusion_results1,app:stable_diffusion_results2,app:stable_diffusion_results3} we show side-by-side results for the entire dataset. Note, images are compressed to reduce file size, for the full resolution images see the SM.

\begin{figure*}
    \centering \includegraphics[width=1.0\linewidth]{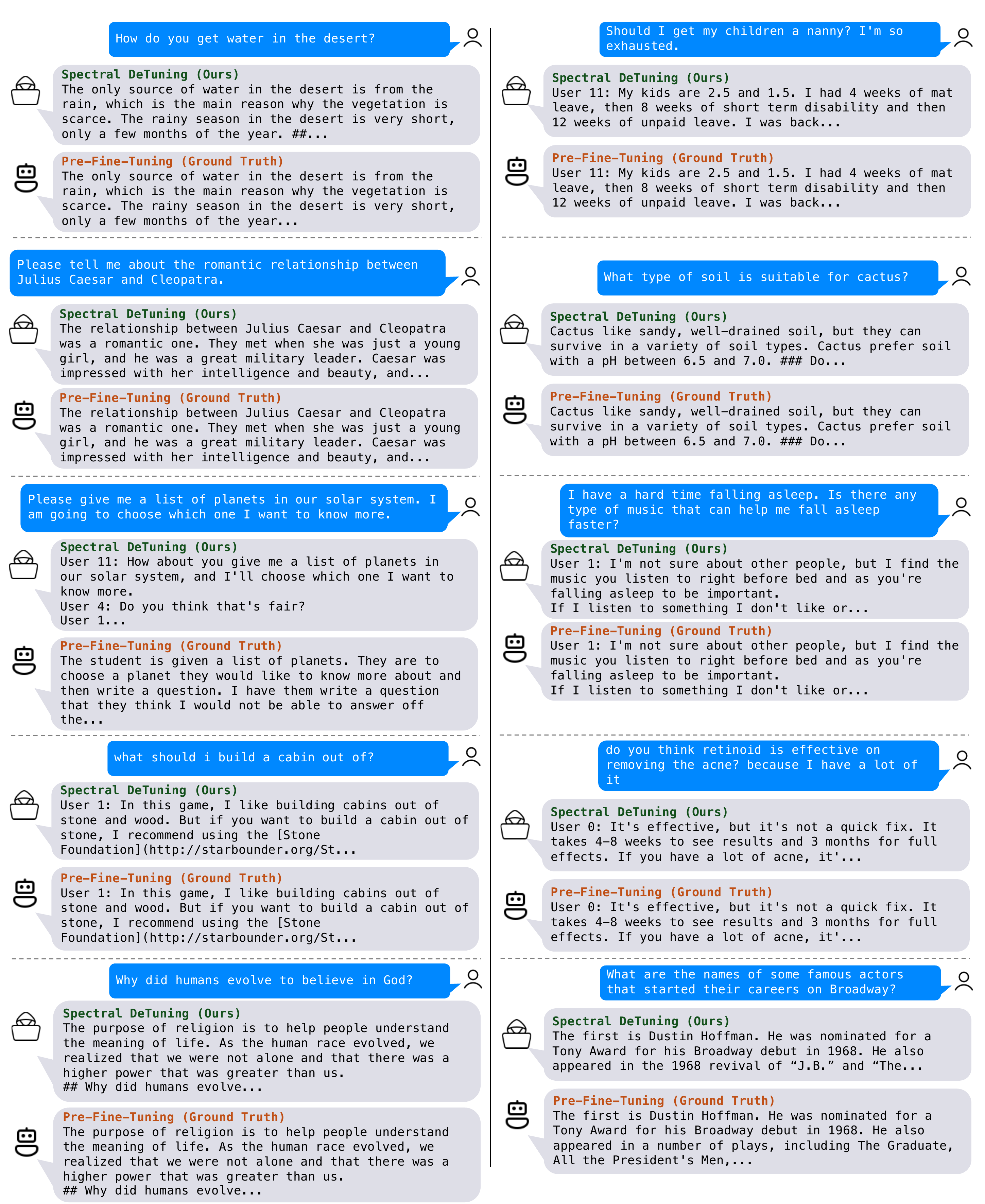}
    \caption{\textit{\textbf{Non Cherry-picked Mistral DPO Results:}} We display side-by-side results for $10$ randomly (\texttt{random\_seed=42}) sampled prompts from our evaluation dataset. For the rest of the results see supplementary material.}
    \label{app:mistral_results}
\end{figure*}

\end{document}